\documentclass[a4paper,fleqn]{cas-sc}

\usepackage[authoryear,longnamesfirst]{natbib}

\usepackage{array}
\usepackage{booktabs}
\usepackage{multirow}
\usepackage{subcaption}
\usepackage{url} 
\usepackage{graphicx}
\usepackage{float} 

\def\tsc#1{\csdef{#1}{\textsc{\lowercase{#1}}\xspace}}
\tsc{WGM}
\tsc{QE}

\begin{document}
\let\WriteBookmarks\relax
\def\floatpagepagefraction{1}
\def\textpagefraction{.001}

\shorttitle{MKMed}    

\shortauthors{Xiang L, Haixu M, Chen L, Shunpan L}  

\title [mode = title]{Combating the Bucket Effect:Multi-Knowledge Alignment for Medication Recommendation}  



%

\author[1]{Xiang Li}
\ead{lixiang_222@stumail.ysu.edu.cn}
\credit{Conceptualization, Methodology, Software, Data Curation, Writing - Original Draft, Visualization, Project administration, Investigation}
\cormark[1]

\author[1]{Haixu Ma}
\ead{mahaixu@stumail.ysu.edu.cn}
\credit{Methodology, Software, Validation, Data Curation, Writing - Original Draft, Writing - Review \& Editing, Visualization}
\cormark[1]
\cortext[1]{First Author and Second Author contribute equally to this work.}

\author[1]{Guanyong Wu}
\ead{wuguanyong@stumail.ysu.edu.cn}
\credit{Software, Validation, Data Curation, Writing - Review \& Editing}

\author[1]{Shi Mu}
\ead{mushi@stumail.ysu.edu.cn}
\credit{Software, Writing - Review \& Editing}

\author[1]{Chen Li}
\ead{lichen36211@gmail.com}
\credit{Writing - Original Draft, Writing - Review \& Editing, Supervision}
\cormark[2]

\author[1,2]{Shunpan Liang}
\ead{liangshunpan@ysu.edu.cn}
\credit{Supervision, Funding acquisition}
\cormark[2]
\cortext[2]{{Corresponding author.}}

\affiliation[1]{
            organization={School of Information Science and Engineering, Yanshan University},
            city={Qin Huangdao},
            postcode={066004}, 
            country={China}
}
\affiliation[2]{
            organization={Xinjiang College Of Science \& Technology},
            city={Korla},
            postcode={841000}, 
            country={China}
}

\begin{abstract}
Medication recommendation is crucial in healthcare, offering effective treatments based on patient's electronic health records (EHR). 
Previous studies show that integrating more medication-related knowledge improves medication representation accuracy. However, not all medications encompass multiple types of knowledge data simultaneously. For instance, some medications provide only textual descriptions without structured data.  This imbalance in data availability limits the performance of existing models, a challenge we term the \textit{“bucket effect”} in medication recommendation. Our data analysis uncovers the severity of the \textit{“bucket effect”} in medication recommendation. 
To fill this gap, we introduce a cross-modal medication encoder capable of seamlessly aligning data from different modalities and propose a medication recommendation framework to integrate \textbf{M}ultiple types of \textbf{K}nowledge, named MKMed.
Specifically,  we first pre-train a cross-modal encoder with contrastive learning on five knowledge modalities, aligning them into a unified space. Then, we combine the multi-knowledge medication representations with patient records for recommendations. Extensive experiments on the MIMIC-III and MIMIC-IV datasets demonstrate that MKMed mitigates the \textit{“bucket effect”} in data, and significantly outperforms state-of-the-art baselines in recommendation accuracy and safety.
\end{abstract}




\maketitle

\section{Introduction}
\label{intro}
Given the increasing demand for healthcare resources, there is a growing emphasis on AI-based medical systems. Medication recommendations~\cite{gamenet,cognet,stratmed}, as a key area, aim to integrate clinical knowledge with patient electronic health records (EHR), enhancing the accuracy, safety, and efficiency of clinical decision-making for patients.

Existing methods can be divided into two categories. 
The first category focuses on exploring the complex relationships between multiple medical events, optimizing patient representation by constructing complex networks\cite{longitudinal1,longitudinal2,longitudinal3}. For example,  RAREMed~\cite{zhao2024leave} focuses on the connections between rare events and others. However, these methods often suffer from inaccurate medication representation, resulting in relationship networks that lack clear medical significance.
The second category of methods~\cite{safedrug,carmen} considers incorporating external medical knowledge, such as molecular structures or textual descriptions, to enhance medication representation. For example, MoleRec~\cite{molerec} analyzes the effects of substructures on patient health. NLA-MMR~\cite{tan2024natural} focuses on textual descriptions of medication functions. These studies demonstrate that introducing more medication-related external medical knowledge helps improve the accuracy of medication recommendation.

\begin{figure}
    \centering
    \includegraphics[width=\textwidth]{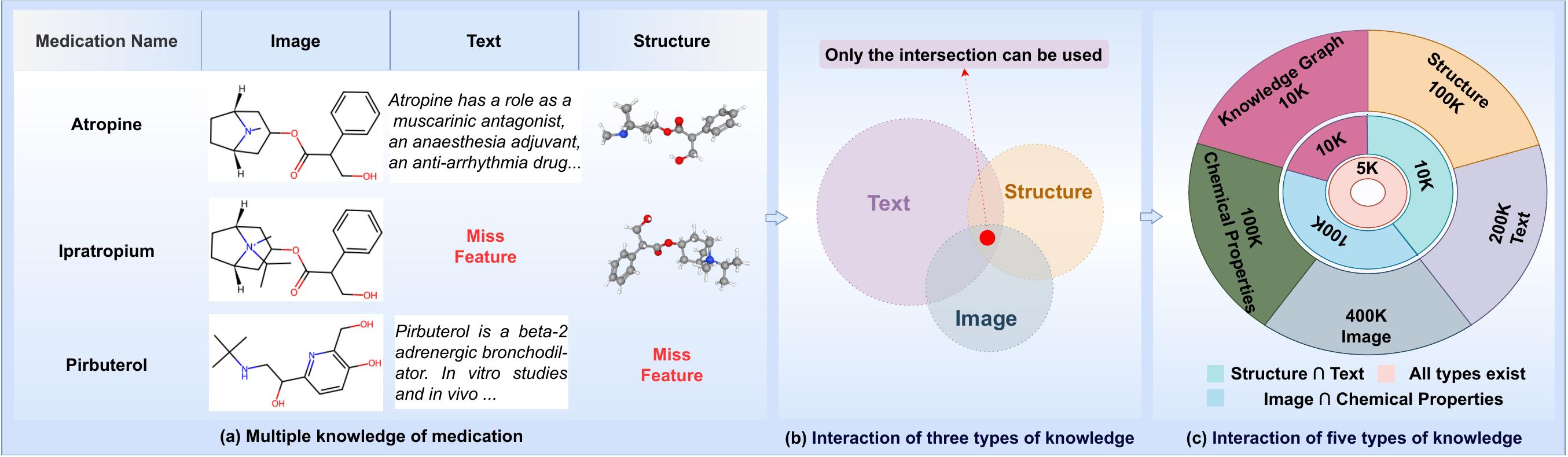}
    \caption{The sparse intersection of multiple data.}
    \label{fig:fig_2}
\end{figure}

However, previous methods \cite{molerec,wu2023megacare} tend to rely on a single type of knowledge, as illustrated in figure~\ref{fig:fig_2} (a). 
Medications like "Atropine" and "Ipratropium" may appear visually similar in their images, but they possess significantly different spatial structures. 
If medication recommendation systems base their decisions solely on the image type, these two medications could potentially be mapped to the same representation, even though they are distinct in terms of their therapeutic uses. 
This limitation highlights the inadequacy of relying solely on a single type of knowledge. 
To fully capture the unique characteristics, complexities, and nuances of each medication, it is essential to integrate multiple types of knowledge. 
By combining different types of information, the system can develop a more comprehensive and holistic understanding of medications. 
This, in turn, enables the generation of more accurate and more nuanced medication representations.
Furthermore, current approaches to multi-knowledge fusion depend on modality alignment, necessitating training on the overlapping components of multi-modal data. 
Specifically, as illustrated in figure~\ref{fig:fig_2} (a), a medication typically does not encompass multiple types of knowledge data simultaneously.
As depicted in figure~\ref{fig:fig_2} (b), only a small proportion of medications include datasets that integrate multiple types of knowledge, highlighting the limited overlap between different knowledge domains. 
However, the types of modalities become more abundant, as shown in figure~\ref{fig:fig_2} (c), the shared or common parts among different types of knowledge become increasingly sparse. 
This phenomenon, in which the intersection of diverse knowledge areas becomes more fragmented as the number of modalities expands, is referred to as the "bucket effect" in this paper, which is especially prominent in medical data.

This issue makes it challenging to apply existing multi-modal methods directly to medication recommendations, as these methods tend to ignore a large amount of valuable information outside the intersection. In other words, the model’s performance is heavily constrained by the sample size of the common part of the modalities. As a result, the model fails to leverage the broader range of information available, leading to considerable information loss. 

To fill this gap, we propose a novel \textbf{M}ulti-\textbf{K}nowledge fusion framework for \textbf{med}ication recommendation, namely MKMed, which is capable of seamlessly aligning data from different modalities based on a cross-modal medication encoder.
Specifically, we first introduce the cross-modal medication encoder, which encodes the molecular structure information contained in each medication. Based on contrastive learning, we align the cross-modal encoder with multiple types of knowledge, enabling the five types of information to complement each other and form a unified representation space. Finally, the pre-trained cross-modal encoder is used to encode medications in the EHR, which are then combined with patient historical records to generate medication recommendations.
Experiments on the MIMIC-III and MIMIC-IV datasets show that MKMed mitigates the information loss and improves the effectiveness. We have released our implementation details on Github\footnote{\url{https://github.com/MKMed-2025/MKMed}}.

In summary, the main contributions of this paper are as follows:
\begin{itemize}
    \item We observe that the performance of existing multi-modal methods is constrained by the sample size of the intersection across multiple data types, and define it as the “bucket effect.” These methods overlook a significant amount of valuable information outside the intersection, leading to severe information loss.
    
    \item To the best of our knowledge, we are the first to propose a framework that integrates multiple types of medication-related knowledge for medication recommendation, significantly enriching medication feature representation.
    
    \item We introduce a cross-modal medication encoder that aligns five types of knowledge into a unified representation space through contrastive learning, without requiring complete co-occurrence of all data types.
    
    \item Experiments on the MIMIC-III and MIMIC-IV datasets show that MKMed mitigates the “bucket effect” and outperforms baseline methods across multiple metrics, highlighting its effectiveness in medication recommendation.
\end{itemize}

\section{Related Work}
\label{subsec1}

In this section, we will introduce the related work from two aspects: medication recommendation and multi-modal molecules.

\subsection{Medication Recommendation}
Mainstream medication recommendation systems can be broadly divided into two main approaches:

The first category is data-driven medication recommendation methods, with early models~\cite{earlywork1,earlywork2,earlywork3} relying on statistical methods. Recently, as deep learning technology has rapidly advanced, many researchers have applied these methods to medication recommendation systems. Initial deep learning models primarily focused on patients' current health status, often overlooking medical history. For instance, LEAP~\cite{leap} formulates medication recommendation as a multi-instance, multi-label sequential decision-making process, leveraging a content-based attention variant of sequence-to-sequence models to extract key features from multiple patient visits.As EHR data has expanded, a shift toward mining patients' diagnostic, procedural, and medication histories to construct personalized recommendations has become mainstream. For example, RETAIN~\cite{retain} models longitudinal patient data with a two-level attention mechanism to focus on crucial past visits for diagnostic and treatment recommendations, though it does not explicitly address recommendation safety. GAMENet~\cite{gamenet} models longitudinal EHR data as a graph structure, incorporating DDI from medication knowledge bases as a DDI graph to mitigate potential conflicts. Similarly, COGNet~\cite{cognet} applies a Transformer-based architecture, using a translation approach to infer medications based on disease and procedure information. StratMed~\cite{stratmed} introduces a dual-attribute network to balance safety and accuracy in medication combinations, enhancing both aspects.

The second category consists of knowledge-driven medication recommendation methods, which improve accuracy in matching medications to patient needs by incorporating external knowledge, such as two-dimensional or three-dimensional molecular structures and substructures. For example, SafeDrug~\cite{safedrug} enhances recommendation precision and quality by using molecular structure embeddings, moving beyond reliance on medication history and reducing the risk of DDIs. Similarly, MoleRec~\cite{molerec} increases accuracy and effectiveness by modeling the relationship between patient health conditions and medication molecular substructures. Carmen~\cite{carmen} goes further by integrating patient medical records into the molecular representation learning process, boosting the system’s ability to distinguish subtle molecular differences.

This knowledge-driven approach to medication recommendation addresses the limitations of data-driven methods in medication representation learning. Integrating molecular-related knowledge has proven to be an effective way to enhance model performance. However, current studies predominantly utilize only the two-dimensional structure modality of molecules, overlooking valuable information in other modalities, such as text descriptions and three-dimensional structures. This limitation restricts the model’s ability to capture comprehensive molecular representations, thereby affecting the performance of downstream tasks.
\subsection{Multi-Modal RSs}
The emergence of RSs aims to address the issue of information overload by analyzing users' historical interaction data and predicting item ratings, thereby recommending the most relevant content to users. However, traditional RSs~\cite{traRSs1,traRSs2,traRSs3,traRSs4} primarily rely on single-modality data and require extensive interactions between users and items to achieve more accurate recommendations. Consequently, multimodal RSs~\cite{mmgcl,dragon,dualgnn} have gained widespread attention in academia and industry in recent years. Their core objective is to leverage information from multiple modalities (e.g., text, images, audio, and video) to provide users with more accurate and personalized recommendations. 

Multimodal data presents both opportunities and challenges. On the positive side, multimodal information enhances input data, allowing for the capture of diverse features~\cite{ducho}. However, describing each modality comprehensively poses challenges, particularly with missing modalities. Traditional methods seek to maximize modality correlation by finding a shared low-dimensional subspace~\cite{privileged, hotelling, deep}, potentially neglecting inter-modality complementarity, which can lead to suboptimal outcomes. More advanced approaches attempt to recover missing modalities by leveraging existing modalities via deep models~\cite{missing_1,visual} or cross-modality recovery strategies~\cite{found,missing_2}. However, these methods often demand substantial amounts of complete multimodal data. Additionally, privileged information learning methods~\cite{aslam,depth,automated}, although capable of leveraging primary and supplementary modalities, necessitate readily accessible training data, limiting their applicability in scenarios with incomplete modalities. This paper proposes a novel approach to mitigate the challenges posed by missing modalities.

\section{Preliminary}
\subsection{Medical Entity}
In this study, the medical entities involved are divided into three categories: diseases, procedures, and medications. Among them, diseases \(\mathcal{D}\) refer to the diagnostic conditions of patients; procedures \(\mathcal{P}\) include various medical operations or treatment procedures; and medications \(\mathcal{M}\) refer to the prescription medications used by patients. These three categories of entities are represented as \(\mathcal{D} = \{d_1, d_2, \ldots\}\), \(\mathcal{P} = \{p_1, p_2, \ldots\}\), and \(\mathcal{M} = \{m_1, m_2, \ldots\}\).

\subsection{Input and Output}
This paper utilizes EHR as the data source, encompassing the entire diagnostic and treatment process of patients. Each patient's EHR is represented as \(\mathcal{H}\), which contains multiple visit records \(\mathcal{H} = \{v_1, v_2, \ldots, v_t\}\), where each visit record \(v_t\) represents the clinical data of the patient during the \(t\text{-th}\) visit. Each visit record \(v_t\) includes three key components: diseases, procedures, and medication information. For each visit \( v_t = \{\mathcal{D}_t, \mathcal{P}_t, \mathcal{M}_t\} \), which includes three types of medical entities, where \( \mathcal{D}_t \subset \mathcal{D} \), \( \mathcal{P}_t \subset \mathcal{P} \), and \( \mathcal{M}_t \subset \mathcal{M} \). For the \( t \)-th visit \( v_t \), we predict the appropriate medication combination \( \hat{\mathcal{M}}_t \) for the patient based on the diagnosis and procedure information \( \mathcal{D}_t \) and \( \mathcal{P}_t \).

\subsection{DDI Matrix}
Drug-Drug Interaction (DDI) refers to the interaction between two or more drugs when used simultaneously, which may cause side effects or affect the treatment's efficacy. Introducing the DDI matrix into a medication recommendation system helps identify and avoid potential harmful medication combinations, ensuring that the recommended medication plan not only achieves therapeutic effects but also reduces the risk of adverse reactions caused by drug interactions, thereby improving medication safety. Specifically, we construct a symmetric matrix \(M\in\{0,1\}\) to represent the interaction relationships between medications. In this matrix, an element \( M_{ij} \) with a value of 1 indicates that there is an interaction between drug \( m_i \) and drug \( m_j \), while a value of 0 indicates no interaction between them.


\subsection{Multimodal Data}
\label{sec:multimodal_data}
This study focuses on integrating molecular data from five modalities to enhance medication molecular representations: Image, Text, Structure, Molecular Properties, and Knowledge Graph. These data modalities provide complementary information from different perspectives of molecules. 
\textbf{Image}: Captures visual representations of molecular structures, enabling the extraction of shape-based features.  
\textbf{Text}: Provides natural language data, enriching semantic understanding.  
\textbf{Structure}: Encodes the spatial arrangement of molecules, capturing interatomic spatial relationships and stereochemical features, and offers essential geometric structural data.  
\textbf{Chemical Properties}: Quantitative descriptors, such as molecular weight,  hydrogen bond acceptor(HBA), hydrogen bond donor(HBD), polar surface area(PSA), aromatic rings(AROM), and other physicochemical properties, are crucial for characterizing molecular behavior.  
\textbf{Knowledge Graph}: Represents a knowledge network of molecules and their biological relationships, capturing their roles and associations within biological systems. This enhances representation from semantic and systems biology perspectives.  

For KG data, we use the DRKG dataset~\cite{drkg} and the TransE method~\cite{transe} to generate the pre-trained model. In addition, all data are matched with their corresponding SMILES\footnote{\url{https://go.drugbank.com/data_packages}} strings, sourced from PubChem~\cite{pubchem}.

\subsection{Molecular Information}
\label{sec:molecular_information}
We use the \textit{Breaking Retrosynthetically Interesting Chemical Substructures} (BRICS) method to decompose molecules into substructures, which is provided as an API in the RDKit~\cite{rdkit} package. Information at the sub-structure and molecule is also converted into graph structures using methods in the same package. 

\section{Methods}
\begin{figure}
    \centering
    \includegraphics[width=\textwidth]{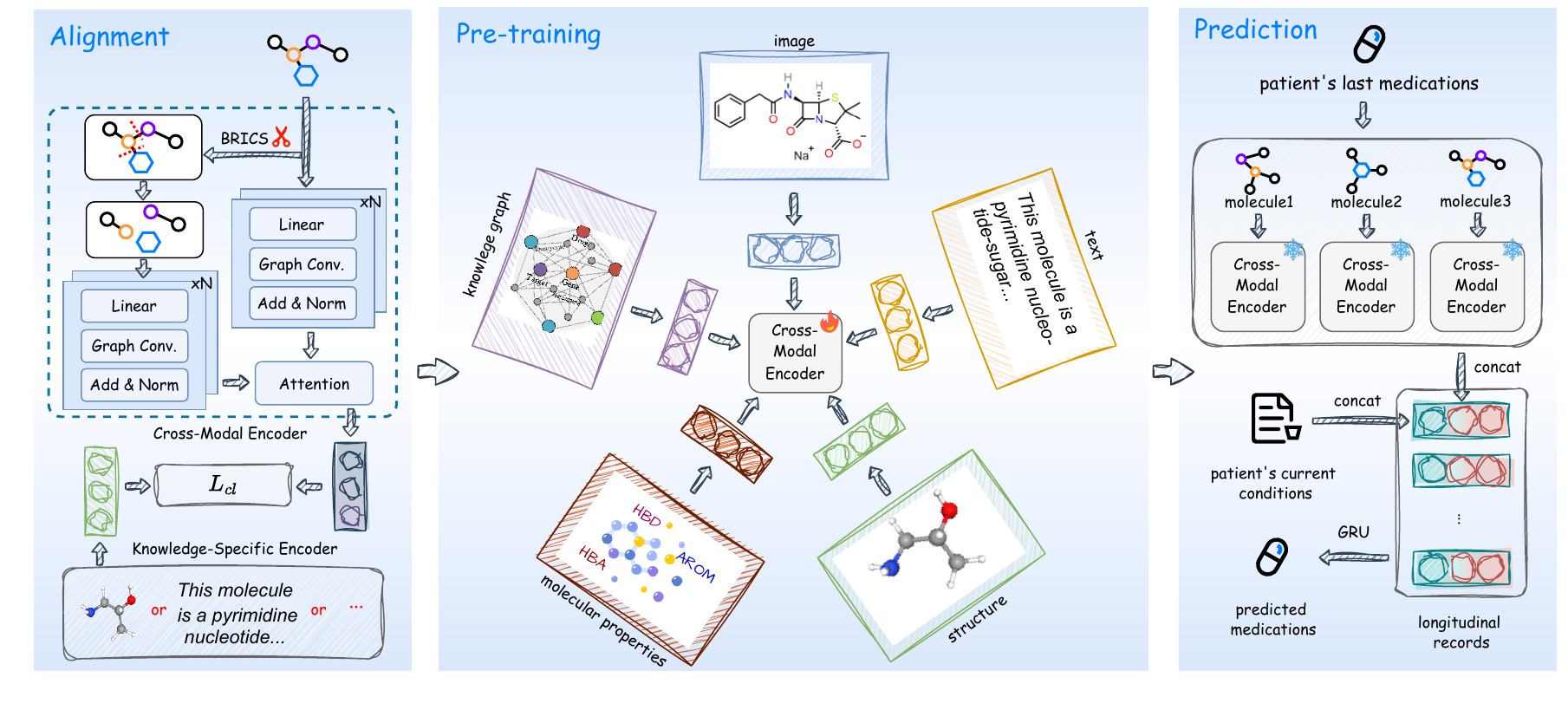}
    \caption{MKMed: Left: Demonstrates the detailed process of knowledge embedding generation and alignment. Mid: The cross-modal encoder pre-training module, which integrates five distinct knowledge sources into the cross-modal encoder. Right: The prediction module, which first utilizes the pretrained encoder to generate medication representations, then combines them with the patient's current condition to produce the final medication recommendation.}
    \vspace{-0.2in}
    \label{fig:model}
\end{figure}

Figure \ref{fig:model} illustrates our model, which is comprised of three main components. 
(1) \textbf{\textit{Modality Alignment}}: This module aims to align embeddings from different representation spaces into a unified space, mirroring the pre-training phase.
(2) \textbf{\textit{Cross-Modal Encoder Pre-training Module}}: This module trains the cross-modal encoder using molecular multi-modal data, employing contrastive learning to integrate structural, text, and image information, capturing relationships across modalities.  
(3) \textbf{\textit{Prediction Module}}: This module leverages the cross-modal encoder from the pre-training phase to create more comprehensive medication representations and generate longitudinal patient representations by incorporating other relevant medical entities. Based on these representations,  recommends medications for the patient.

\subsection{Modality Alignment}
The input to this module consists of the molecular SMILES string \( C_i \) and its corresponding observation \( O_i \) in another modality \( O \). First, we encode them into embedding representations: \( \mathbf{e}_{C_i} = \mathbf{f}(C_i) \), \( \mathbf{e}_{O_i} = \mathbf{g}(O_i) \), where \( \mathbf{f} \) represents our cross-modal encoder and \( \mathbf{g} \) represents the modality-specific encoder. After obtaining the embeddings for each modality, to address the issue of different modality features being distributed in separate representation spaces, we introduce a contrastive learning mechanism to align the features of each modality into a shared representation space.

Specifically, given the embedding matrix \( \mathbf{E}_O = [\mathbf{e}_{O_1}, \mathbf{e}_{O_2}, \dots, \mathbf{e}_{O_N}] \) and its paired embedding matrix \( \mathbf{E}_C = [\mathbf{e}_{C_1}, \mathbf{e}_{C_2}, \dots, \mathbf{e}_{C_N}] \), we calculate the cosine similarity matrix between the two sets of features to measure their degree of alignment. Then, through a learnable scaling factor, we adjust the values in the similarity matrix to enhance the contrastive signal between positive and negative samples. Finally, we align the two sets of features into the same representation space by computing the contrastive loss in both directions:
\begin{gather}
\mathrm{L}(\mathrm{\mathbf{E}_{C},\mathbf{E}_{O}})=-\frac{1}{N}\sum_{i=1}^{N}\left[\log\frac{e^{\tau \cdot sim(\mathbf{e}_{C_i},\mathbf{e}_{O_i})}}{\sum_{j= 1}^{N}e^{\tau \cdot sim(\mathbf{e}_{C_i},\mathbf{e}_{O_j})}}\right], \\
sim(\mathbf{E}_{C}, \mathbf{E}_{O}) = \mathbf{E}_{C} \cdot \mathbf{E}_{O}/(\|\mathbf{E}_{C}\|\|\mathbf{E}_{O}\|),
\end{gather}
where \( \text{sim}(\cdot) \) denotes the cosine similarity, and \( \tau \) is a learnable scaling factor used to adjust the similarity scale between the two vectors.

\subsection{Cross-Modal Encoder Pre-training Module}
To improve the performance of the cross-modal encoder in generating medication molecule representations and reduce model complexity, we introduce a pre-training module that trains the encoder using molecular multi-modal data.
\subsubsection{Cross-Modal Encoder}
Graph Isomorphism Networks (GINs)~\cite{gin} are an ideal choice for handling molecular graphs due to their strong discriminatory power over graphs and their message-passing mechanism, which adapts to both local and global molecular properties. GINs have been widely used in recent medication recommendations and molecular-related research. As with previous work, we use a two-layer GIN to handle graph structure data in the cross-modal encoder. The GIN architecture for both molecular graphs and substructure graphs is the same, but they do not share parameters. The specific iterative process is as follows:
\begin{gather}
    \mathbf{a}_\nu^{(l)}=\mathrm{AGG}^{(l)}(\{\mathbf{h}_\nu^{(l-1)},\mathbf{h}_u^{(l-1)},\mathbf{f}(u,\nu)|u\in\mathcal{N}(\nu)\}),\\
    \mathbf{h}_{\nu}^{(l)}=\mathrm{COMBINE}^{(l)}(\mathbf{h}_{\nu}^{(l-1)},\mathbf{a}_{\nu}^{(l)}),
\end{gather}
here, \(\mathcal{N}(\nu)\) represents the neighbors of node \(\nu\), \(u\) is a neighbor, and \(\mathbf{h}_\nu^{(l)}\) is the node feature vector at layer \(l\). \(\mathbf{f}(u, \nu)\) denotes edge features, while \(\mathrm{AGG}^{(l)}\) aggregates node and edge features from layer \(l\)-1. \(\mathrm{COMBINE}^{(l)}\) merges the aggregated vector \(\mathbf{a}_\nu^{(l)}\) with the features from the previous layer.

Taking the molecule graph \( G \) of a molecule as an example, after \( l \) layers of the above operations, we summarize the feature representations of all nodes through a readout function to obtain the molecule embedding representation of the molecule:
\begin{equation}
    \mathbf{h}_{G}=\mathrm{READOUT}(\mathbf{h}_{\nu}^{(L)}|\nu\in V).
\end{equation}

All molecule representations are aggregated into an embedding matrix \( \mathbf{E}_m \). Similarly, for the substructure, a substructure embedding matrix \( \mathbf{E}_{sub} \) is constructed. Since different substructures have varying levels of importance to the overall molecule, we fuse these embeddings at different levels based on an attention mechanism: 
\begin{gather}
    \mathbf{A}_{(mol, sub)}(\mathrm{X,Y})=\mathrm{Softmax}\left(\mathrm{mask}\left(\frac{\mathbf{Q}_x\mathbf{K}_y^\top}{\sqrt{d}}\right)\right),\\
    \mathbf{e}_{m_i}=\mathbf{A}_{(mol, sub)}\mathbf{E}_{sub_i},
\end{gather}
where \(d\) represents the maximum embedding dimension in \( X \) and \( Y \), and the mask function replaces the missing values in the matrix with the minimum value. Finally, the embeddings aggregated from different levels of information are integrated into an embedding matrix \( \mathbf{E}_C \).

\subsubsection{Modality-Specific Encoders}
\textbf{Image Encoder. }For the image modality, we use the CLIP Vision Transformer (ViT)~\cite{vit} to process molecular images. ViT divides images into fixed-size patches, captures molecular microstructures, and uses self-attention to aggregate patch information into a global representation, effectively modelling long-range dependencies.

Specifically,for each input image \( I \), iterative process is as follows
\begin{gather}
    \mathbf{e}_j = \text{flatten}(I_{j}) \mathbf{W}_{img},\\
    \mathbf{e}_I = \mathrm{ViT}(\mathbf{e}_1,\mathbf{e}_2,...\mathbf{e}_n),
\end{gather}
here, \( I_j \) represents the \( j \)-th patch of \( I \), and \(\mathbf{W}_{img} \) is a learnable weight matrix.
Afterward, all the image representations are combined into the image embedding matrix \( \mathbf{E}_{img} \), where each row corresponds to the embedding of a molecular image.

\noindent\textbf{Text Encoder.} For the text modality, we use the CLIP text encoder~\cite{clip} to generate high-dimensional, semantically rich embeddings. Input text is truncated if necessary, tokenized into sequences \( T = \{ t_1, t_2, \dots, t_n \} \), and positional encodings (\( p_i \)) are added to form \( T' = \{ t_1 + p_1, t_2 + p_2, \dots, t_n + p_n \} \). The enhanced sequence \( T' \) is processed by a Transformer, and for multi-segment texts, segment embeddings are summed to produce the final representation.
\begin{equation}
    \mathbf{e}_{t} = \text{Text Encoder}(T').
\end{equation}

This process generates a molecular text description embedding matrix \( \mathbf{E}_{t} \), where each row represents the embedding of a molecular description. 

\noindent\textbf{KG Encoder.} For the KG modality, we follow the pre-training model derivation process outlined in Section~\ref{sec:multimodal_data} to construct the embedding table \( \mathbf{E}_{kg} \).

\noindent\textbf{Chemical Properties Encoder.} For the chemical properties modality, we first use the RDKit API to obtain the chemical properties of each molecule and then extract the property features using the CLIP-based tokenization method. Finally, we generate the embedding for this modality through a linear layer and aggregate them into the embedding table \( \mathbf{E}_{mp} \).

\noindent\textbf{Structure Encoder.} For the structure modality, we use a Geometric Vector Perceiver (GVP)~\cite{gvp} to capture high-dimensional molecular information in geometric space. GVP processes scalar (\( \mathbf{s} \)) and vector (\( \mathbf{V} \)) features while preserving spatial relationships, preserving the spatial relationships of the molecule and preventing information loss. By applying convolution layers, GVP transforms node and edge features to represent the molecular structure effectively. The specific transformation process is as follows:
\begin{gather}
    \mathbf{f}_{sv} = \mathrm{CONCAT}(\|\mathbf{W}_h\mathbf{V}\|_{2,\mathrm{row-wise}},\mathbf{s}),\\
    \mathbf{f}_v = \mathbf{W}_{\mu}\mathbf{W}_h\mathbf{V},\\
    \mathbf{s}^{\prime} = \mathrm{MLP}_1(\mathbf{f}_{sv}),\\
    \mathbf{V}^{\prime} = \sigma(\|\mathbf{f}_v\|_{2,\mathrm{row-wise}})\odot \mathbf{f}_v.
\end{gather}

After processing through the GVP layer, a global pooling operation is applied to aggregate all node representations into a scalar representation of the input molecule, resulting in the final structure embedding matrix \( \mathbf{E}_{S} \), where each row represents the embedding of a molecule.

\subsection{Prediction Module}
\subsubsection{Patient representation learning}
Using longitudinal electronic medical records, we encode a patient's health status based on diseases, procedures, and medications. Learnable embedding tables \( \mathbf{E}_d \) and \( \mathbf{E}_p \) store embeddings for diseases and procedures, mapping multi-hot vectors of the patient's current conditions into an embedding space. Meanwhile, for the patient's current medication representation, we use the cross-modal encoder derived during the pre-training phase to generate the medication embedding representation:
\begin{gather}            
    \mathbf{e}_{d_{t}}=d_t\mathbf{E}_{d},\mathbf{e}_{p_{t}}=p_t\mathbf{E}_{p},\\
    \mathbf{e}_{m_{t}} = \mathrm{f}(m_t),
\end{gather}
here, \(\mathrm{f}\) represents the cross-modal encoder. We then model the sequences of diseases, procedures, and medications using GRU to capture the temporal dependency information of the current patient:
\begin{equation}
    \mathbf{h}_{x_{i}} = \mathrm{GRU}_{x}(\mathbf{e}_{x_{i}}, \mathbf{h}_{x_{i-1}}), \quad x \in \{d, p, m\},
\end{equation}

where \( \mathbf{h}_{x_{i}} \) denotes the hidden state, and \( \mathrm{GRU}_{x} \) represents the corresponding GRU model (e.g., \( \mathrm{GRU}_{d} \), \( \mathrm{GRU}_{p} \), \( \mathrm{GRU}_{m} \)), initialized as zero vectors and serving as the basis for patient representations. Finally, these states are flattened and concatenated to form a comprehensive patient representation for prediction tasks:
\begin{equation}
     \mathbf{e}_i = \text{CONCAT}(\mathbf{h}_{d_{i}}, \mathbf{h}_{p_{i}}, \mathbf{h}_{m_{i}}). 
\end{equation}

\subsubsection{Prediction}
Given the final patient representation as input, we generate prediction scores using an MLP. Then, by setting a threshold \(\delta\), we select entries that exceed this threshold to obtain the recommended medication combinations.
\begin{gather}
    \mathbf{e}_m = \mathrm{MLP}_2(\mathbf{e}_i),\\
    \hat{m}_i=\begin{cases} 1,&\mathrm{if}\quad \mathbf{e}_m(i)\geq\delta\\ 0,&\mathrm{if}\quad \mathbf{e}_m(i)<\delta\end{cases},
\end{gather}
here, \(\hat{m}_i\) is a multi-hot vector representing predicted medications, where \(i\) denotes the \(i\)-th medication in \(\mathbf{e}_m\). A value of \(m=1\) indicates the medication is recommended, and \(m=0\) means it is not.

\subsection{Model Training}
In this study, we design a deep learning-based medication recommendation system aimed at generating safe and effective medication combinations while minimizing potential drug-drug interactions (DDI). Moreover, the medication recommendation task can essentially be viewed as a multi-label binary classification problem. Given the specificity of medication recommendations, we employ three primary loss functions to optimize the model: binary cross-entropy loss \(\mathcal{L}_{bce}\), multi-label margin loss \(\mathcal{L}_{multi}\), and DDI loss \(\mathcal{L}_{ddi}\)~\cite{molerec}. The definitions of these three functions are as follows:

\textbf{Binary Cross-Entropy Loss (BCE).}
The binary cross-entropy (BCE) loss is used to optimize the model's correct predictions of drug combinations. For each patient's medication combination recommendation problem, we treat it as a multi-label classification task, where each label corresponds to a specific medication. The BCE loss is defined as follows:
\begin{equation}
    \mathcal{L}_{bce} = -\sum_{i=1}^{|\mathcal{M}|} m_i \log(\hat{m}_i) + (1 - m_i) \log(1 - \hat{m}_i)
\end{equation}

\textbf{Multi-Label hinge loss.}
To further enhance the model's ability to correctly predict multiple medication labels, we introduce the multi-label hinge loss. This loss function ensures that the model can effectively distinguish between medications by maximizing the score difference between actual medication labels and other negative samples. For the multi-label prediction problem of a sample, the loss function is defined as:
\begin{equation}
    \mathcal{L}_{multi} = \sum_{i,j: m_i=1, m_j=0} \frac{\max(0, 1 - (\hat{m}_i - \hat{m}_j))}{|\mathcal{M}|}
\end{equation}

\textbf{DDI loss.}
We introduce the DDI loss \( \mathcal{L}_{\text{ddi}} \)~\cite{molerec} to constrain the recommended medication combinations, aiming to avoid including too many high-risk drug-drug interactions. The loss is defined as:
\begin{equation}
    \mathcal{L}_{ddi} = \sum_{i=1}^{|\mathcal{M}|} \sum_{j=1}^{|\mathcal{M}|} M_{ij} \cdot \hat{m}_i \cdot \hat{m}_j
\end{equation}

\textbf{Combined Controllable Loss Function}

The total loss function of the model is a weighted combination of the three loss functions described above:
\begin{equation}
    \mathcal{L} = \beta(\gamma {\mathcal{L}_{bce}} + (1 - \gamma) {\mathcal{L}_{multi}}) + (1 - \beta) {\mathcal{L}_{ddi}}
\end{equation}
where \( \beta \) and \( \gamma \) are hyperparameters. This combination allows us to optimize the accuracy of medication recommendations while also considering the safety between medications, ensuring that the output medication combinations have a lower risk of DDI.

\section{Experiments}
In this section, we conduct extensive experiments on our proposed MKMed, aiming to answer the following research questions (RQs):
\begin{itemize}
    \item \textbf{RQ1}: Does MKMed significantly outperform other state-of-the-art models' performance?
    \item \textbf{RQ2}: Do the key modules proposed in this paper enhance the model’s performance?
    \item \textbf{RQ3}: How does the number of modalities affect the model's performance? 
    \item \textbf{RQ4}: How do different modality alignment methods compare in terms of their impact on performance? 
    \item \textbf{RQ5}: How sensitive is the model's performance to changes in its hyperparameters?
\end{itemize}  

\subsection{Setup Protocol}
We will first introduce the specific experimental environment, the configuration and parameters of the model, as well as the sampling methods used during the testing phase.

\subsubsection{Experimental Environment}
All experiments are conducted on a machine running Ubuntu 22.04, equip-ped with 30 GB of memory, 12 CPUs, a 24GB NVIDIA RTX 3090 GPU, PyTorch 2.0.0, and CUDA 11.7.

\subsubsection{Configuration and Parameter}
For the entity embedding tables \( E_d \), \( E_p \), and \( E_m \), we use 64 as the embedding size with an initialization range of -0.1 to 0.1. The molecular graph neural network (GIN) has 2 layers, each with a hidden dimension of 64. The initial node features for each molecular graph are 9-dimensional, covering atom count, chirality information, and other atomic properties. The 3D edge features include bond type, stereochemistry, and conjugation. The spatial edge features include bond type, stereochemistry, and conjugation. The overall molecular structure features consist of molecular conformation, spatial coordinates, radial basis function (RBF) distances, and normalized inter-atomic vectors. Additionally, we use a 3-layer GVP model, where the scalar hidden dimension of node features is 128, the vector hidden dimension is 64, and the scalar and vector hidden dimensions of edge features are 32 and 1, respectively. The gated recurrent unit (GRU) employs 64 hidden units.For the loss functions, the same hyperparameters are used during training, testing, and validation: threshold \( \delta = 0.5 \), \( \beta = 0.95 \), and a DDI acceptance rate of 0.06. In the pretraining stage, we run for 20 epochs, while the formal training stage runs for 25 epochs. The model is trained using the Adam optimizer, with a learning rate of 1e-6 for pretraining and 5e-4 for formal training, and a regularization factor \( R = 0.05 \).

\subsubsection{Sampling Approach}
Due to the limited availability of publicly accessible EHR data, we employed a bootstrapping sampling method at this stage, following the recommendation in SafeDrug~\cite{safedrug}. This technique is particularly suited for scenarios with small sample sizes, as discussed in the relevant literature, making it highly effective for addressing data scarcity issues.

\subsection{Datasets}

\begin{table}
    \caption{Statistics of the datasets.}
    \centering
    \begin{tabular}{|c|c|c|}
    \toprule
    Item & MIMIC-III & MIMIC-IV \\
    \midrule
    \# patients         & 6,350  & 60,125 \\
    \# clinical events  & 15,032 & 156,810 \\
    \# diseases         & 1,958  & 2,000 \\
    \# procedures       & 1,430  & 1,500 \\
    \# medications      & 131   & 131\\
    avg. \# of visits    & 2.37  & 2.61\\
    avg. \# of medications & 11.44  & 6.66\\
    \bottomrule
    \end{tabular}
    \label{tab:datasets}
\end{table}

MIMIC-III~\cite{mimic3} and MIMIC-IV~\cite{mimic4} are two widely used clinical datasets for medical research, containing detailed EHR of intensive care unit (ICU) patients, including demographics, diagnoses, treatments, and laboratory results. The MIMIC-III dataset contains 15,032 visits, 6,350 patients, 131 medications, 1,958 diagnoses, and 1,430 procedures. The MIMIC-IV dataset includes 156,810 visits, 60,125 patients, 131 medications, 2,000 diagnoses, and 1,500 procedures. We split the datasets into training, validation, and test sets in a 2/3, 1/6, 1/6 ratio and detailed the statistical information of the processed data, as shown in the Table~\ref{tab:datasets}.

\subsection{Evaluation Metrics}

We employ four commonly used metrics in medication recommendation—\\Jaccard, DDI rate, F1-score, and PRAUC to thoroughly analyze and evaluate the performance of our approach. In the following sections, we will describe in detail how each metric is calculated and how it is applied in our study.

\textbf{Jaccard} (Jaccard similarity score) is a statistical metric used to measure the similarity between two sets. It evaluates their similarity by calculating the ratio of the intersection to the union of the two sets. A higher Jaccard score indicates a greater consistency between the predicted medication combinations and the actual medications, reflecting higher accuracy.
\begin{gather}
    Jaccard(t) = \frac{|\{i:\hat{m}_i=1\}|\cap|\{i:{m}_i=1\}|}{|\{i:\hat{m}_i=1\}|\cup |\{i:{m}_i=1\}|}, \\
    Jaccard = \frac{1}{N_h}\sum_{t=1}^{N_h} Jaccard(t),
\end{gather}
where $\hat{m}_i$ represents the multi-hot vector of the predicted outcome, $m_i$ represents the real label, $Jaccard(t)$ represents the evaluation result at visit $t$, and $N_h$ represents the total number of visits for patient $h$.

\textbf{DDI} (Drug-Drug Interaction rate) is an important criterion for assessing the safety of medication combinations, reflecting the probability of potential interactions between medications. A lower DDI rate indicates that the recommended medication combinations are safer and more effective when used.
\begin{equation}
    DDI = \frac{\sum_{i=1}^{N_h}\sum_{k,l\in \{j:\hat{m}_j(t)=1\}}1\{a^{ddi}_{kl}=1\}}{\sum_{i=1}^{N_h}\sum_{k,l\in \{ j:m_j(t)=1\}}1 },
\end{equation}
where \(N_h\) denotes the total number of visits for patient \(h\), \(m(t)\) and \(\hat{m}(t)\) denote the real and predicted multi-labels at visit \(t\), \(m_j(t)\) denotes the \(j^{th}\) entry of \(m(t)\), \(a^{ddi}\) is the prior DDI relation matrix, and \(1\) is an indicator function that returns 1 when \(a^{ddi} = 1\), otherwise 0.

\textbf{F1} (F1-score) is a metric used to comprehensively evaluate model performance by combining precision and recall. In medication recommendation, a high F1 score indicates that the model is not only able to accurately recommend the appropriate medications but also effectively captures all potential medication choices.
\begin{gather}
    Precision(t) = \frac{|\{i:\hat{m}_i=1 \}\cap \{i:m_i=1\}|}{|\{i:\hat{m}_i=1\}|},\\
    Recall(t) = \frac{|\{i:\hat{m}_i=1 \}\cap \{i:m_i=1\}|}{|\{i:m_i=1\}|},\\
    F1(t) = \frac{2}{\frac{1}{Precision(t)}+\frac{1}{Recall(t)}},\\
    F1 = \frac{1}{N_h}\sum_{i=1}^{N_h}F1(i),
\end{gather}
where $N_h$ represents the total number of visits for patient $h$.

\textbf{PRAUC} (Precision-Recall Area Under Curve) represents the area under the precision-recall curve and is used to evaluate model performance. A higher PRAUC score indicates that the model can maintain good precision at higher recall rates, meaning it recommends correct medication combinations while keeping the error rate low.
\begin{gather}
    PRAUC(t) = \sum_{k=1}^{|M|} Precision_{k}(t)\triangle Recall_k(t),\\
    \triangle Recall_k(t) = Recall_k(t) - Recall_{k-1}(t),
\end{gather}
where $|M|$ denotes the number of medications, $k$ is the rank in the sequence of the retrieved medications, and $Precision_{k}(t)$ represents the precision at cut-of $k$ in the ordered retrieval list and $\triangle Recall_k(t)$ denotes the change of recall from medication $k-1$ to $k$. We averaged the PRAUC across all of the patient's visits to obtain the final result,
\begin{equation}
    PRAUC = \frac{1}{N_h}\sum_{i=1}^{N_h}PRAUC(t),
\end{equation}
where $N_h$ represents the total number of visits for patient $h$.

\textbf{Avg. \# Med} (Average number of medications) measuring the number of medications in each recommendation, reflect the complexity of the regimen. A higher value may increase treatment complexity, while a lower value suggests that the medication regimen may be more streamlined and manageable, helping to reduce unnecessary medication use. However, this metric serves only as a reference, and the appropriateness of the medication combination should still be evaluated based on the clinical context.
\begin{equation}
    Avg.\# Med = \frac{1}{N_h}\sum_{i=1}^{N_h}|\hat{M}(i)|,
\end{equation}
where $N_h$ represents the total number of visits for patient $h$ and $|\hat{M}(i)|)$ denotes the number of predicted medications in visit $i$ of patient $h$.

\subsection{Baselines}
We select the following representative state-of-the-art methods as baselines to test our model:

\textbf{RETAIN}~\cite{retain} is an attention model for analyzing medical sequential data. It captures key events in a patient's medical history by combining temporal and feature information, thereby providing more accurate personalized analyses for disease prediction and diagnosis.

\textbf{LEAP}~\cite{leap} enhances the effectiveness of medications by simulating label dependencies through the decomposition of treatment processes into sequential procedures.

\textbf{GAMENet}~\cite{gamenet} is a medication recommendation model that effectively captures important information and temporal relationships in patient histories by integrating the characteristics of graph neural networks and memory networks. The model leverages graph structures and long-term memory to enhance the accuracy and effectiveness of medication recommendations in medical contexts.

\textbf{SafeDrug}~\cite{safedrug} combines patient health information with molecular knowledge of medications, effectively reducing the impact of drug-drug interactions (DDIs) and recommending safer medication combinations.

\textbf{MICRON}~\cite{micron} recommends medications based on real-time changes in patient health status, dynamically updating the patient's history of medication combinations to adapt to emerging symptoms, thereby maximizing treatment efficacy and minimizing adverse reactions.

\textbf{COGNet}~\cite{cognet} employs a Transformer-based architecture for medication recommendation, utilizing a translation-like approach to infer suitable medications from diseases. It introduces a replication mechanism that integrates effective medications from historical prescriptions into new recommendations, enhancing personalization and accuracy in the medication suggestion process.

\textbf{MoleRec}~\cite{molerec} accurately predicts the mechanisms of action and interactions of medications by simulating the relationships between specific substructures of medication molecules and patient health conditions. By utilizing detailed molecular-level analyses, it offers more personalized and safer medication recommendation solutions.

\textbf{CausalMed}~\cite{causalmed} identifies causal relationships between medical entities through causal discovery, accounting for dynamic differences under varying health conditions and translating these into causally linked medication recommendations.

\subsection{Performance Comparison (RQ1)}

\begin{table*}
\centering
\label{tab:methods_performance}
\caption{The performance of each model on the test set regarding accuracy and safety. The best and the runner-up results are highlighted in bold and underlined respectively under t-tests, at the level of 95\% confidence level.}
\setlength{\extrarowheight}{1.05pt}
\resizebox{\textwidth}{!}{
\begin{tabular}{l@{}l|c|c|c|c|c|c@{}}
\toprule
 & Datasets & Method & Jaccard$\uparrow$ & DDI$\downarrow$ & F1-Score$\uparrow$ & PRAUC$\uparrow$ & Avg.\# Med. \\ 
\midrule

 &  & RETAIN & 0.4871 ± 0.0038 & 0.0879 ± 0.0012 & 0.6473 ± 0.0033 & 0.7600 ± 0.0035 & 19.4222 ± 0.1682 \\
 &  & LEAP & 0.4526 ± 0.0035 & 0.0762 ± 0.0006 & 0.6147 ± 0.0036 & 0.6555 ± 0.0035 & 18.6240 ± 0.0680 \\
 &  & GAMENet & 0.4994 ± 0.0033 & 0.0890 ± 0.0005 & 0.6560 ± 0.0031 & 0.7656 ± 0.0046 & 27.7703 ± 0.1726 \\
 &   & SafeDrug & 0.5126 ± 0.0027 & 0.0587 ± 0.0005 & 0.6692 ± 0.0025 & 0.7660 ± 0.0022 & 18.9837 ± 0.1187 \\
 & MIMIC-III & MICRON & 0.5219 ± 0.0021 & \underline{0.0727 ± 0.0009} & 0.6761 ± 0.0018 & 0.7489 ± 0.0034 & 19.2505 ± 0.2618 \\
 &  & COGNet & 0.5316 ± 0.0020 & 0.0858 ± 0.0008 & 0.6644 ± 0.0018 & 0.7707 ± 0.0021 & 27.6279 ± 0.0802 \\
 &  & MoleRec & 0.5301 ± 0.0025 & 0.0756 ± 0.0006 & 0.6841 ± 0.0022 & 0.7748 ± 0.0022 & 22.2239 ± 0.1661 \\
 &  & CausalMed & \underline{0.5389 ± 0.0011} & \textbf{0.0709 ± 0.0007} & \underline{0.6916 ± 0.0022} & \underline{0.7826 ± 0.0010} & 20.5419 ± 0.1645\\
 \rowcolor[HTML]{F5F5F5}&  & \textbf{MKMed} & \textbf{0.5526* ± 0.0022}   & 0.0740 ± 0.0005  & \textbf{0.7036* ± 0.0019}   & \textbf{0.7961* ± 0.0021}    & 22.7991 ± 0.1624  \\ 
\midrule
 &  & RETAIN & 0.4234 ± 0.0017 & 0.0936 ± 0.0015 & 0.5785 ± 0.0013 & 0.6801 ± 0.0012 & 10.9576 ± 0.0818\\ 
 &  & LEAP & 0.4254 ± 0.0013 & 0.0688 ± 0.0011 & 0.5794 ± 0.0017 & 0.6059 ± 0.0018 & 11.3606 ± 0.0812\\
 &  & GAMENet & 0.4565 ± 0.0018 & 0.0898 ± 0.0012 & 0.6103 ± 0.0019 & 0.6829 ± 0.0017 & 18.5895 ± 0.0815\\
 &   & SafeDrug & 0.4487 ± 0.0012	& \textbf{0.0604 ± 0.0010} & 0.6014 ± 0.0017 & 0.6948 ± 0.0018 & 13.6943 ± 0.0714\\
 & MIMIC-IV & MICRON & 0.4640 ± 0.0017 & 0.0691 ± 0.0015 & 0.6167 ± 0.0016 & 0.6919 ± 0.0014 & 12.7701 ± 0.0711\\
 &  & COGNet & 0.4775 ± 0.0014 & 0.0911 ± 0.0013 & 0.6233 ± 0.0019 & 0.6524 ± 0.0018 & 18.7235 ± 0.0715\\
 &  & MoleRec & 0.4744 ± 0.0013 & 0.0722 ± 0.0014 & 0.6262 ± 0.0018	& 0.7124 ± 0.0017 & 13.4806 ± 0.0615\\
 &  & CausalMed & \underline{0.4899 ± 0.0014} & \underline{0.0677 ± 0.0017} & \underline{0.6412 ± 0.0013} & \underline{0.7338 ± 0.0019} & 14.4295 ± 0.0612\\
 \rowcolor[HTML]{F5F5F5}&  & \textbf{MKMed} & \textbf{0.4943* ± 0.0016} & 0.0736 ± 0.0004 & \textbf{0.6461* ± 0.0013} & \textbf{0.7425* ± 0.0019} & 14.5801 ± 0.0627\\ 
\bottomrule
\end{tabular}
}
\label{tab:comparison}
\end{table*}

In this section, we evaluate the accuracy and safety of our model compared to the baseline models. For the baseline methods that provide their models for testing, we directly use the models they provided for evaluation. For those baseline methods that do not provide testing models, we retrain the models and test them, using the optimal parameter settings mentioned in their respective papers to obtain the results. The comparison results are detailed in the Table~\ref{tab:comparison}.

To facilitate a more intuitive comparison of the experimental results, we summarize all evaluation metrics into a separate chart. This helps to clearly illustrate the comparison between the baseline models and our method. Additionally, we recorded efficiency metrics such as the number of epochs required for convergence, the training time per epoch, and the total training time for each model. Since different models show significant variance in efficiency, efficiency-related metrics are not included in the table. The table contains Jaccard, F1-score, and PRAUC to evaluate model accuracy, while DDI and Avg.\# Med are used to assess model safety.

First, LEAP, a subsequently proposed deep learning method that treats medication recommendation as a generative task, shows lower effectiveness compared to other traditional models. Although RETAIN adopts a sequence-based recommendation approach, its DDI rate remains relatively high.

Next, methods like GAMENet started incorporating patient history information into the model, significantly improving accuracy, but also leading to higher DDI rates. SafeDrug successfully reduced DDI rates by explicitly modeling medication molecular structures, but it fell short in fully addressing the complex relationships between medications and diseases, leaving room for accuracy improvements. MICRON, by dynamically updating the patient’s medication history to accommodate new symptoms, greatly enhanced model accuracy while also demonstrating fast training speeds. COGNet used a translation-model-like approach with a copy mechanism, improving personalization and accuracy of recommendations. MoleRec introduced external molecular knowledge, modeling the connection between molecular substructures and diseases, achieving significant improvements across metrics without notably affecting safety. However, this method has not yet accounted for the potential impact of multimodal molecular information on model performance.

Our model integrates external knowledge from multiple molecular modalities, without requiring all pairwise combinations of these modalities. This significantly improves the model's ability to leverage information, addressing the limitations posed by low information utilization. By doing so, our approach achieves a substantial boost in accuracy without notably compromising safety, ultimately ensuring a well-balanced trade-off between accuracy and safety in medication recommendations.

\section{Discussions}
In this section, we conduct an in-depth analysis of the aforementioned experimental results to highlight the advancements of this framework. Additionally, we further demonstrate the effectiveness of the framework through supplementary experiments.

\subsection{Effectiveness Analysis}
The above experimental results indicate that our model shows improvement in accuracy; therefore, we conduct a more in-depth analysis of the experimental results.

Early medication recommendation studies enhance the performance of recommendation systems by incorporating the time-dependent information of patient visits. Subsequent research introduces external molecular knowledge, further improving the accuracy and safety of the system. However, existing studies mainly focus on single-modal external knowledge, neglecting the multi-source domain knowledge of medication molecules. This oversight leads to certain limitations in the model's ability to learn medication molecular representations, resulting in suboptimal recommendation outcomes. Moreover, previous multimodal models exhibit insufficient data utilization, relying solely on the intersection of multiple modal data.

To address these issues, we incorporate the multimodal information of medication molecules into the model and design a rotating self-supervised pre-training module. This module integrates multimodal information into the molecular representation, not only supplementing the missing multimodal information in the model but also enhancing data utilization. Ultimately, the proposed model effectively resolves the issues overlooked by the baseline model, significantly improving the overall performance of the medication recommendation system.

\begin{table}
\centering
\caption{The performance of each ablation model on the test set regarding accuracy and safety. The best and the runner-up results are highlighted
in bold and underlined respectively under t-tests, at the level of 95\% confidence level.}
\begin{tabular}{@{}|l|c|c|c|c|}
\toprule
Datasets & \multicolumn{4}{c|}{MIMIC-III/ MIMIC-IV}\\
\midrule
Measures& Jaccard$\uparrow$&F1-Score$\uparrow$ &PRAUC$\uparrow$&DDI$\downarrow$\\
\midrule
\textbf{MKMed}$_{mol}$ &0.5357/ 0.4781 &0.6819/ 0.6314 &0.7815/ 0.7267 &0.0728/ 0.0730\\
\textbf{MKMed}$_{pt}$  &0.5381/ 0.4834 &0.6894/ 0.6320 &0.7853/ 0.7331 &0.0743/ 0.0735\\
\textbf{MKMed}$_{pm}$  &\underline{0.5435/ 0.4861} &\underline{0.6953/ 0.6374} &\underline{0.7868/ 0.7338} &0.0745/ 0.0740\\
\midrule
\textbf{MKMed}         &\textbf{0.5526/ 0.4943} &\textbf{0.7036/ 0.6461} &\textbf{0.7961/ 0.7425} &0.0740/ 0.0736\\
\bottomrule
\end{tabular}
\label{tab:ablation}
\end{table}

\subsection{Efficiency Analysis}
To assess the model's efficiency, we first decompose the model and analyze the time complexity of each module. Our model consists of a multimodal pre-training module, a patient representation module, and a medication recommendation module. In the pre-training module, the two-dimensional structure encoder and the three-dimensional spatial structure encoder utilize GIN and GVP graph neural networks, respectively, with a time complexity of \(O((N+E)^2)\), where \(N\) represents the number of nodes, \(E\) denotes the number of edges, and 2 is the number of GNN layers. For the molecular image encoder and text description encoder, we employ a Transformer-based architecture, resulting in a time complexity of \(O(L \cdot (X^2 \cdot \text{dim} + X \cdot \text{dim}^2))\), where \(L\) is the number of Transformer layers, \(X\) is the number of patches (or the length of the text sequence), and \(\text{dim}\) is the embedding dimension of patches (or tokens). Therefore, the time complexity of the pre training module is:
\begin{equation}
    O_{pre}( 2 \cdot (N+E)^2 + 2 \cdot (L \cdot (X^2 \cdot \text{dim} + X \cdot \text{dim}^2)) )
\end{equation}
In the patient representation module, we use an embedding layer, which has a time complexity of \(O(L \cdot \text{dim})\), where \(L\) is the sequence length and \(\text{dim}\) is the embedding dimension. The medication recommendation module consists of several gated recurrent units (GRU) and linear layers, with the time complexity of the GRU layer being \(O(g \cdot h^2)\), where \(g\) represents the sequence length and \(h\) is the size of the hidden layer. Therefore, the overall time complexity of the model can be expressed as:
\begin{equation}
O( O_{pre} + (N+E)^2 + g \cdot h^2 + L \cdot \text{dim})
\end{equation}
\begin{figure}
    \centering
    \includegraphics[width=0.8\linewidth]{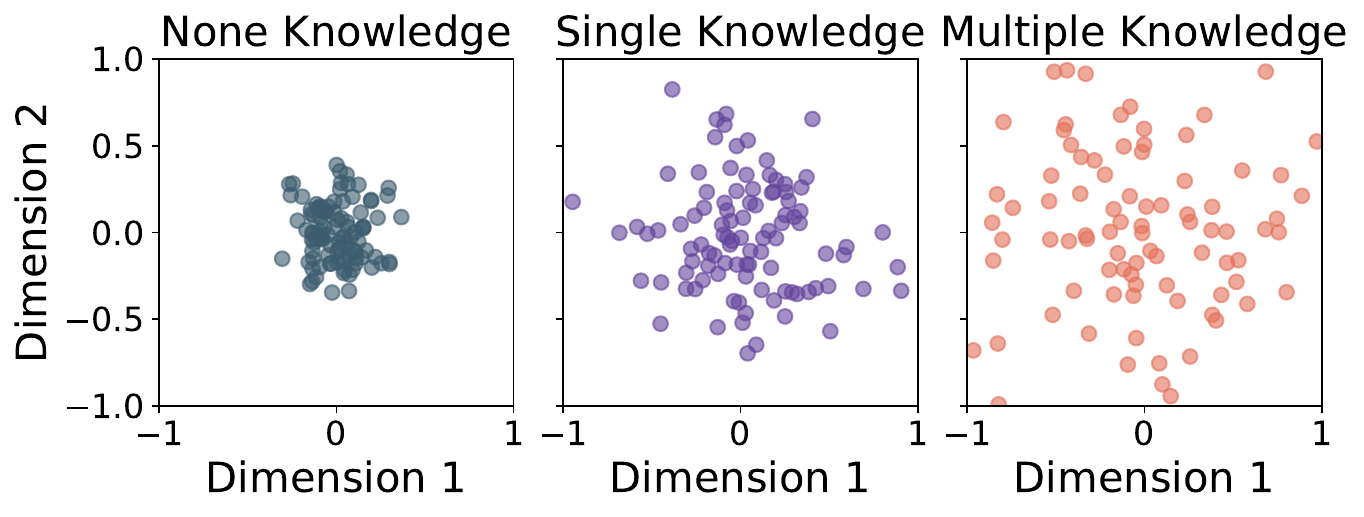}
    \caption{The low-dimensional data distribution represented by molecular embeddings. From left to right: No external knowledge, using one type of knowledge, and using all five types of knowledge.}
    \label{fig:tsne}
\end{figure}
\subsection{Ablation Study (RQ2)}
Specifically, we design three variants: \textbf{MKMed}$_{mol}$ that omits molecular representation and instead uses raw medication data; \textbf{MKMed}$_{pt}$ that removes the pre-training module and uses an untrained cross-modal encoder; and \textbf{MKMed}$_{pm}$ that trains using only the structural modality data. 

The ablation results in Table~\ref{tab:ablation} firstly confirm that each component plays a vital role. Secondly, removing the molecular representation (\textbf{MKMed}$_{mol}$) leads to a significant performance drop, highlighting the importance of molecular representation in capturing chemical and biological properties essential for understanding medication characteristics. Next, \textbf{MKMed}$_{pt}$ indicates that the critical role of pretraining in enriching representation space and enhancing feature extraction for molecular embeddings. Lastly, \textbf{MKMed}$_{pm}$ demonstrates that while using a single knowledge limits model performance, partial modality data still offers useful features compared to ignoring multi-knowledge information entirely.

\subsection{Modality Number and Performance (RQ3)}
We first analyze the cross-modal encoder trained with different numbers of modalities using the t-SNE method to examine how varying modalities affect feature representations. As shown in figure \ref{fig:tsne}, the results reveal clear trends: when the encoder, which has not been trained, is used to construct feature representations, the resulting representations are more concentrated in the feature space, indicating that the encoder has not yet fully learned the feature information of each modality. As more modality data is introduced, the feature representations constructed by the encoder gradually become more dispersed. This phenomenon suggests that, with the increase in modalities, the encoder is able to progressively learn and integrate diverse feature information from different modalities. This capability of multimodal learning enables the encoder to better distinguish between different molecules in molecular representation, thereby enhancing the overall performance of the model.

\begin{figure}
    \centering
    \includegraphics[width=0.8\linewidth]{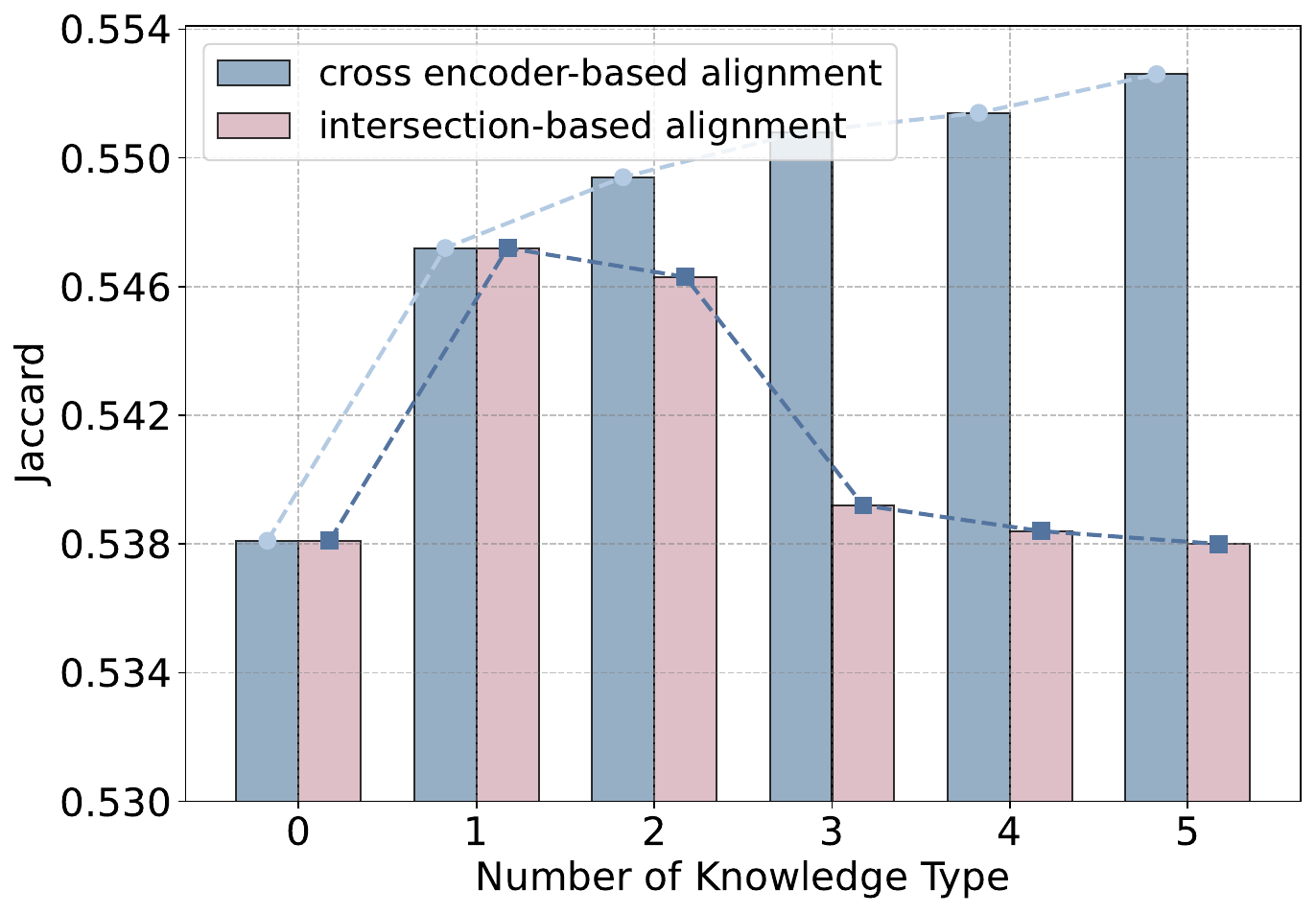}
    \caption{Differences in the performance changes of different alignment methods when the number of modaliteis changes.}
    \label{fig:auxi_combined}
\end{figure}

Next, to visually demonstrate how the quantity of knowledge modalities affects model performance, we employ the Jaccard metric and construct six experimental configurations corresponding to different knowledge-type quantities (0 to 5 types). As shown by the left column in figure~\ref{fig:auxi_combined}, the results reveal a consistent performance enhancement as more knowledge types are incorporated. This improvement occurs because combining diverse knowledge sources enables richer feature representations. The multi-knowledge approach demonstrates clear superiority over configurations using only single knowledge types.

\subsection{Modality data alignment comparison (RQ4)}

In this experiment, we focus on using only the intersecting portion of the multi-modal data as input to the pre-training module. The intersecting portion refers to the shared data points across all knowledge types, ensuring that the model is trained on data that is common to all the knowledge types, and the experimental results presented on the right column in figure~\ref{fig:auxi_combined}.  This phenomenon can be attributed to the inverse relationship between the number of knowledge types and the amount of intersecting data. As the number of knowledge types increases, the intersection of data between them becomes smaller, meaning that the model is trained on a more limited dataset. Notably, the model’s performance does not improve with the addition of more knowledge types; instead, it starts to decline after reaching three knowledge types. This decline can be explained by the fact that the performance of the model is largely constrained by the knowledge type with the smallest sample size. As additional knowledge types are introduced, the available intersecting data becomes more sparse, limiting the model’s ability to learn from sufficient data, which ultimately affects its performance. 

\begin{figure}
    \centering
    \begin{subfigure}{\linewidth}
        \centering
        \includegraphics[width=\linewidth]{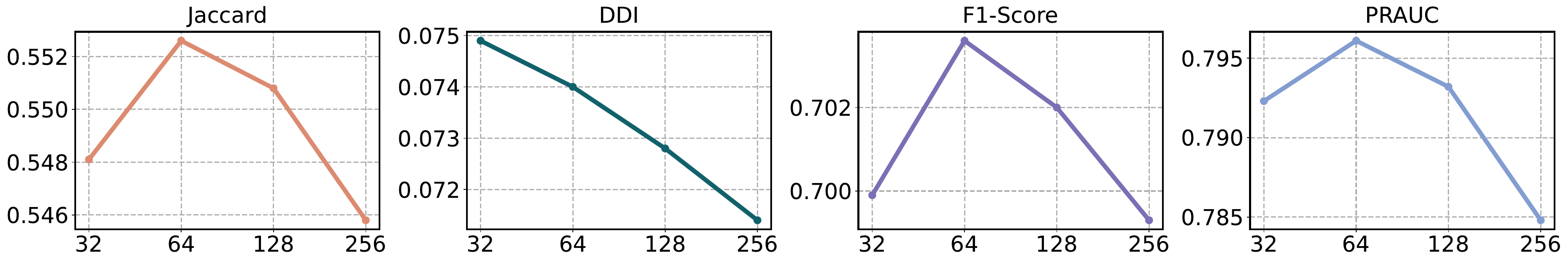}
        \caption{The impact of $dim$.}
        \label{fig:param_dim}
    \end{subfigure}
    \vfill
    \begin{subfigure}{\linewidth}
        \centering
        \includegraphics[width=\linewidth]{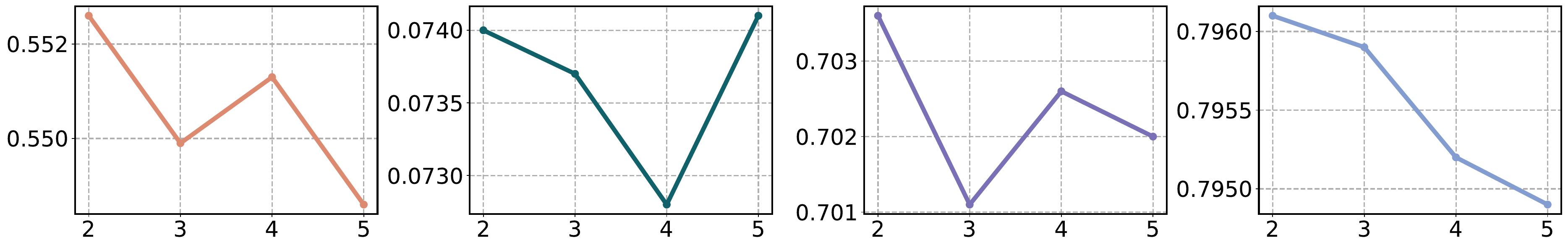}
        \caption{The impact of $GNN \ layer$.}
        \label{fig:param_layer}
    \end{subfigure}
    \caption{Experiments on parameter sensitivity on MIMIC-III dataset.}
    \label{fig:params}
\end{figure}

\subsection{Parameter Sensitivity (RQ5)}

To evaluate the robustness and sensitivity of the model under different hyperparameter settings, we conducted a comprehensive parameter sensitivity analysis. This experiment focused on key hyperparameters, including embedding dimensions and the number of GNN layers, by training and validating the model under various parameter combinations. By analyzing the performance variations across these combinations, we identified parameters that significantly impact model performance and further fine-tuned them to improve results. In this experiment, we used Jaccard, F1-score, PRAUC, and DDI rate to quantify model performance and visualized the influence of different hyperparameters on the model through line charts.

We first investigated the impact of embedding dimensions on model performance. As shown in Figure \ref{fig:param_dim}, we evaluated the model's performance across embedding dimensions ranging from 32 to 256. The results indicate that as the embedding dimension increased, the model's performance improved significantly in the early stages but declined after the dimension exceeded 64. This may be due to excessively high dimensions leading to a sharp increase in the number of model parameters, introducing redundant representations and raising the risk of overfitting. Additionally, higher dimensions incurred significant computational overhead. At an embedding dimension of 64, our model achieved a good balance between accuracy and efficiency, ensuring both high predictive performance and reasonable computational resource consumption. Therefore, we set the embedding dimension to 64 as the final model configuration.

Next, we evaluated the impact of the number of graph neural network (GNN) layers on model performance. We tested network architectures with 2 to 5 layers, and the experimental results are shown in Figure \ref{fig:param_layer}. Overall, the model exhibited optimal accuracy with 2 GNN layers, and further increasing the network depth did not improve performance but instead led to a noticeable decline. This phenomenon may occur because more layers introduce additional trainable parameters, making the model more prone to overfitting when training data is limited or feature noise is present, thereby widening the gap between training and testing performance. Moreover, from a practical application perspective, increasing the number of layers also significantly raised computational resource demands and reduced inference efficiency. Therefore, after considering performance, generalization capability, and computational cost, we ultimately selected a 2-layer GNN as the model's foundational structure.

In practical model experiments, flexible adjustments should be made based on the experimental environment and actual results. Different experimental conditions (e.g., hardware configurations, dataset characteristics, and training frameworks) can influence model training outcomes.

\section{Conclusion}
The MKMed framework proposed in this paper introduces an innovative alignment strategy that effectively integrates multiple knowledge, addressing the performance of existing models is limited by the size of the intersection of multiple types of knowledge data. Compared to existing baseline methods, MKMed aligns five different types of medical knowledge, image, structure, text, chemical properties, and knowledge graphs into a unified representation space through a contrastive learning pre-training module, significantly improving the accuracy and comprehensiveness of medication representation. Experimental results demonstrate that MKMed outperforms the latest benchmark methods on publicly available clinical datasets, proving its effectiveness and broad application potential in the field of medication recommendation.

\printcredits

\bibliographystyle{cas-model2-names}

\bibliography{MKMed}

\begin{thebibliography}{47}
\expandafter\ifx\csname natexlab\endcsname\relax\def\natexlab#1{#1}\fi
\providecommand{\url}[1]{\texttt{#1}}
\providecommand{\href}[2]{#2}
\providecommand{\path}[1]{#1}
\providecommand{\DOIprefix}{doi:}
\providecommand{\ArXivprefix}{arXiv:}
\providecommand{\URLprefix}{URL: }
\providecommand{\Pubmedprefix}{pmid:}
\providecommand{\doi}[1]{\href{http://dx.doi.org/#1}{\path{#1}}}
\providecommand{\Pubmed}[1]{\href{pmid:#1}{\path{#1}}}
\providecommand{\bibinfo}[2]{#2}
\ifx\xfnm\relax \def\xfnm[#1]{\unskip,\space#1}\fi
\bibitem[{An et~al.(2021)An, Zhang, You, Tian, Jin and Wei}]{earlywork2}
\bibinfo{author}{An, Y.}, \bibinfo{author}{Zhang, L.}, \bibinfo{author}{You, M.}, \bibinfo{author}{Tian, X.}, \bibinfo{author}{Jin, B.}, \bibinfo{author}{Wei, X.}, \bibinfo{year}{2021}.
\newblock \bibinfo{title}{Mesin: Multilevel selective and interactive network for medication recommendation}.
\newblock \bibinfo{journal}{Knowledge-Based Systems} \bibinfo{volume}{233}, \bibinfo{pages}{107534}.
\bibitem[{Aslam et~al.(2023a)Aslam, Zeeshan, Pedersoli, Koerich, Bacon and Granger}]{privileged}
\bibinfo{author}{Aslam, M.H.}, \bibinfo{author}{Zeeshan, M.O.}, \bibinfo{author}{Pedersoli, M.}, \bibinfo{author}{Koerich, A.L.}, \bibinfo{author}{Bacon, S.}, \bibinfo{author}{Granger, E.}, \bibinfo{year}{2023}a.
\newblock \bibinfo{title}{Privileged knowledge distillation for dimensional emotion recognition in the wild}, in: \bibinfo{booktitle}{Proceedings of the IEEE/CVF conference on computer vision and pattern recognition}, pp. \bibinfo{pages}{3338--3347}.
\bibitem[{Aslam et~al.(2023b)Aslam, Zeeshan, Pedersoli, Koerich, Bacon and Granger}]{aslam}
\bibinfo{author}{Aslam, M.H.}, \bibinfo{author}{Zeeshan, M.O.}, \bibinfo{author}{Pedersoli, M.}, \bibinfo{author}{Koerich, A.L.}, \bibinfo{author}{Bacon, S.}, \bibinfo{author}{Granger, E.}, \bibinfo{year}{2023}b.
\newblock \bibinfo{title}{Privileged knowledge distillation for dimensional emotion recognition in the wild}, in: \bibinfo{booktitle}{Proceedings of the IEEE/CVF conference on computer vision and pattern recognition}, pp. \bibinfo{pages}{3338--3347}.
\bibitem[{Attimonelli et~al.(2024)Attimonelli, Danese, Malitesta, Pomo, Gassi and Di~Noia}]{ducho}
\bibinfo{author}{Attimonelli, M.}, \bibinfo{author}{Danese, D.}, \bibinfo{author}{Malitesta, D.}, \bibinfo{author}{Pomo, C.}, \bibinfo{author}{Gassi, G.}, \bibinfo{author}{Di~Noia, T.}, \bibinfo{year}{2024}.
\newblock \bibinfo{title}{Ducho 2.0: Towards a more up-to-date unified framework for the extraction of multimodal features in recommendation}, in: \bibinfo{booktitle}{Companion Proceedings of the ACM on Web Conference 2024}, pp. \bibinfo{pages}{1075--1078}.
\bibitem[{Bordes et~al.(2013)Bordes, Usunier, Garcia-Duran, Weston and Yakhnenko}]{transe}
\bibinfo{author}{Bordes, A.}, \bibinfo{author}{Usunier, N.}, \bibinfo{author}{Garcia-Duran, A.}, \bibinfo{author}{Weston, J.}, \bibinfo{author}{Yakhnenko, O.}, \bibinfo{year}{2013}.
\newblock \bibinfo{title}{Translating embeddings for modeling multi-relational data}.
\newblock \bibinfo{journal}{Advances in neural information processing systems} \bibinfo{volume}{26}.
\bibitem[{Chen et~al.(2021)Chen, Niu and Zhang}]{depth}
\bibinfo{author}{Chen, J.}, \bibinfo{author}{Niu, L.}, \bibinfo{author}{Zhang, L.}, \bibinfo{year}{2021}.
\newblock \bibinfo{title}{Depth privileged scene recognition via dual attention hallucination}.
\newblock \bibinfo{journal}{IEEE Transactions on Image Processing} \bibinfo{volume}{30}, \bibinfo{pages}{9164--9178}.
\bibitem[{Chen et~al.(2023)Chen, Li, Geng and Wang}]{carmen}
\bibinfo{author}{Chen, Q.}, \bibinfo{author}{Li, X.}, \bibinfo{author}{Geng, K.}, \bibinfo{author}{Wang, M.}, \bibinfo{year}{2023}.
\newblock \bibinfo{title}{Context-aware safe medication recommendations with molecular graph and ddi graph embedding}, in: \bibinfo{booktitle}{Proceedings of the AAAI Conference on Artificial Intelligence}, pp. \bibinfo{pages}{7053--7060}.
\bibitem[{Choi et~al.(2016)Choi, Bahadori, Sun, Kulas, Schuetz and Stewart}]{retain}
\bibinfo{author}{Choi, E.}, \bibinfo{author}{Bahadori, M.T.}, \bibinfo{author}{Sun, J.}, \bibinfo{author}{Kulas, J.}, \bibinfo{author}{Schuetz, A.}, \bibinfo{author}{Stewart, W.}, \bibinfo{year}{2016}.
\newblock \bibinfo{title}{Retain: An interpretable predictive model for healthcare using reverse time attention mechanism}.
\newblock \bibinfo{journal}{Advances in Neural Information Processing Systems} \bibinfo{volume}{29}.
\bibitem[{Cui et~al.(2022)Cui, Zhou, Wang, Xu and Yang}]{visual}
\bibinfo{author}{Cui, Z.}, \bibinfo{author}{Zhou, L.}, \bibinfo{author}{Wang, C.}, \bibinfo{author}{Xu, C.}, \bibinfo{author}{Yang, J.}, \bibinfo{year}{2022}.
\newblock \bibinfo{title}{Visual micro-pattern propagation}.
\newblock \bibinfo{journal}{IEEE Transactions on Pattern Analysis and Machine Intelligence} \bibinfo{volume}{45}, \bibinfo{pages}{1267--1286}.
\bibitem[{Dosovitskiy(2020)}]{vit}
\bibinfo{author}{Dosovitskiy, A.}, \bibinfo{year}{2020}.
\newblock \bibinfo{title}{An image is worth 16x16 words: Transformers for image recognition at scale}.
\newblock \bibinfo{journal}{arXiv preprint arXiv:2010.11929} .
\bibitem[{He and Cao(2018)}]{automated}
\bibinfo{author}{He, L.}, \bibinfo{author}{Cao, C.}, \bibinfo{year}{2018}.
\newblock \bibinfo{title}{Automated depression analysis using convolutional neural networks from speech}.
\newblock \bibinfo{journal}{Journal of biomedical informatics} \bibinfo{volume}{83}, \bibinfo{pages}{103--111}.
\bibitem[{He et~al.(2020)He, Deng, Wang, Li, Zhang and Wang}]{traRSs1}
\bibinfo{author}{He, X.}, \bibinfo{author}{Deng, K.}, \bibinfo{author}{Wang, X.}, \bibinfo{author}{Li, Y.}, \bibinfo{author}{Zhang, Y.}, \bibinfo{author}{Wang, M.}, \bibinfo{year}{2020}.
\newblock \bibinfo{title}{Lightgcn: Simplifying and powering graph convolution network for recommendation}, in: \bibinfo{booktitle}{Proceedings of the 43rd International ACM SIGIR conference on research and development in Information Retrieval}, pp. \bibinfo{pages}{639--648}.
\bibitem[{Hotelling(1992)}]{hotelling}
\bibinfo{author}{Hotelling, H.}, \bibinfo{year}{1992}.
\newblock \bibinfo{title}{Relations between two sets of variates}, in: \bibinfo{booktitle}{Breakthroughs in statistics: methodology and distribution}. \bibinfo{publisher}{Springer}, pp. \bibinfo{pages}{162--190}.
\bibitem[{Ioannidis et~al.(2020)Ioannidis, Song, Manchanda, Li, Pan, Zheng, Ning, Zeng and Karypis}]{drkg}
\bibinfo{author}{Ioannidis, V.N.}, \bibinfo{author}{Song, X.}, \bibinfo{author}{Manchanda, S.}, \bibinfo{author}{Li, M.}, \bibinfo{author}{Pan, X.}, \bibinfo{author}{Zheng, D.}, \bibinfo{author}{Ning, X.}, \bibinfo{author}{Zeng, X.}, \bibinfo{author}{Karypis, G.}, \bibinfo{year}{2020}.
\newblock \bibinfo{title}{Drkg - drug repurposing knowledge graph for covid-19}.
\newblock \bibinfo{howpublished}{\url{https://github.com/gnn4dr/DRKG/}}.
\bibitem[{Jin et~al.(2018)Jin, Yang, Sun, Liu, Qu and Tong}]{longitudinal2}
\bibinfo{author}{Jin, B.}, \bibinfo{author}{Yang, H.}, \bibinfo{author}{Sun, L.}, \bibinfo{author}{Liu, C.}, \bibinfo{author}{Qu, Y.}, \bibinfo{author}{Tong, J.}, \bibinfo{year}{2018}.
\newblock \bibinfo{title}{A treatment engine by predicting next-period prescriptions}, in: \bibinfo{booktitle}{Proceedings of the 24th ACM SIGKDD International Conference on Knowledge Discovery \& Data Mining}, pp. \bibinfo{pages}{1608--1616}.
\bibitem[{Jing et~al.(2020)Jing, Eismann, Suriana, Townshend and Dror}]{gvp}
\bibinfo{author}{Jing, B.}, \bibinfo{author}{Eismann, S.}, \bibinfo{author}{Suriana, P.}, \bibinfo{author}{Townshend, R.}, \bibinfo{author}{Dror, R.}, \bibinfo{year}{2020}.
\newblock \bibinfo{title}{Learning from protein structure with geometric vector perceptrons}.
\newblock \bibinfo{journal}{Learning,Learning} .
\bibitem[{Johnson et~al.(2023)Johnson, Bulgarelli, Shen, Gayles, Shammout, Horng, Pollard, Hao, Moody, Gow et~al.}]{mimic4}
\bibinfo{author}{Johnson, A.E.}, \bibinfo{author}{Bulgarelli, L.}, \bibinfo{author}{Shen, L.}, \bibinfo{author}{Gayles, A.}, \bibinfo{author}{Shammout, A.}, \bibinfo{author}{Horng, S.}, \bibinfo{author}{Pollard, T.J.}, \bibinfo{author}{Hao, S.}, \bibinfo{author}{Moody, B.}, \bibinfo{author}{Gow, B.}, et~al., \bibinfo{year}{2023}.
\newblock \bibinfo{title}{Mimic-iv, a freely accessible electronic health record dataset}.
\newblock \bibinfo{journal}{Scientific data} \bibinfo{volume}{10}, \bibinfo{pages}{1}.
\bibitem[{Johnson et~al.(2016)Johnson, Pollard, Shen, Lehman, Feng, Ghassemi, Moody, Szolovits, Anthony~Celi and Mark}]{mimic3}
\bibinfo{author}{Johnson, A.E.}, \bibinfo{author}{Pollard, T.J.}, \bibinfo{author}{Shen, L.}, \bibinfo{author}{Lehman, L.w.H.}, \bibinfo{author}{Feng, M.}, \bibinfo{author}{Ghassemi, M.}, \bibinfo{author}{Moody, B.}, \bibinfo{author}{Szolovits, P.}, \bibinfo{author}{Anthony~Celi, L.}, \bibinfo{author}{Mark, R.G.}, \bibinfo{year}{2016}.
\newblock \bibinfo{title}{Mimic-iii, a freely accessible critical care database}.
\newblock \bibinfo{journal}{Scientific data} \bibinfo{volume}{3}, \bibinfo{pages}{1--9}.
\bibitem[{Kim et~al.(2021)Kim, Chen, Cheng, Gindulyte, He, He, Li, Shoemaker, Thiessen, Yu et~al.}]{pubchem}
\bibinfo{author}{Kim, S.}, \bibinfo{author}{Chen, J.}, \bibinfo{author}{Cheng, T.}, \bibinfo{author}{Gindulyte, A.}, \bibinfo{author}{He, J.}, \bibinfo{author}{He, S.}, \bibinfo{author}{Li, Q.}, \bibinfo{author}{Shoemaker, B.A.}, \bibinfo{author}{Thiessen, P.A.}, \bibinfo{author}{Yu, B.}, et~al., \bibinfo{year}{2021}.
\newblock \bibinfo{title}{Pubchem in 2021: new data content and improved web interfaces}.
\newblock \bibinfo{journal}{Nucleic acids research} \bibinfo{volume}{49}, \bibinfo{pages}{D1388--D1395}.
\bibitem[{Landrum(2013)}]{rdkit}
\bibinfo{author}{Landrum, G.}, \bibinfo{year}{2013}.
\newblock \bibinfo{title}{Rdkit documentation}.
\newblock \bibinfo{journal}{Release} \bibinfo{volume}{1}, \bibinfo{pages}{4}.
\bibitem[{Le et~al.(2018)Le, Tran and Venkatesh}]{longitudinal1}
\bibinfo{author}{Le, H.}, \bibinfo{author}{Tran, T.}, \bibinfo{author}{Venkatesh, S.}, \bibinfo{year}{2018}.
\newblock \bibinfo{title}{Dual memory neural computer for asynchronous two-view sequential learning}, in: \bibinfo{booktitle}{Proceedings of the 24th ACM SIGKDD International Conference on Knowledge Discovery \& Data Mining}, pp. \bibinfo{pages}{1637--1645}.
\bibitem[{Li et~al.(2024a)Li, Liang, Hou and Ma}]{stratmed}
\bibinfo{author}{Li, X.}, \bibinfo{author}{Liang, S.}, \bibinfo{author}{Hou, Y.}, \bibinfo{author}{Ma, T.}, \bibinfo{year}{2024}a.
\newblock \bibinfo{title}{Stratmed: Relevance stratification between biomedical entities for sparsity on medication recommendation}.
\newblock \bibinfo{journal}{Knowledge-Based Systems} \bibinfo{volume}{284}, \bibinfo{pages}{111239}.
\bibitem[{Li et~al.(2024b)Li, Liang, Lei, Li, Hou, Zheng and Ma}]{causalmed}
\bibinfo{author}{Li, X.}, \bibinfo{author}{Liang, S.}, \bibinfo{author}{Lei, Y.}, \bibinfo{author}{Li, C.}, \bibinfo{author}{Hou, Y.}, \bibinfo{author}{Zheng, D.}, \bibinfo{author}{Ma, T.}, \bibinfo{year}{2024}b.
\newblock \bibinfo{title}{Causalmed: Causality-based personalized medication recommendation centered on patient health state}, in: \bibinfo{booktitle}{Proceedings of the 33rd ACM International Conference on Information and Knowledge Management}, pp. \bibinfo{pages}{1276--1285}.
\bibitem[{Pham et~al.(2019)Pham, Liang, Manzini, Morency and P{\'o}czos}]{found}
\bibinfo{author}{Pham, H.}, \bibinfo{author}{Liang, P.P.}, \bibinfo{author}{Manzini, T.}, \bibinfo{author}{Morency, L.P.}, \bibinfo{author}{P{\'o}czos, B.}, \bibinfo{year}{2019}.
\newblock \bibinfo{title}{Found in translation: Learning robust joint representations by cyclic translations between modalities}, in: \bibinfo{booktitle}{Proceedings of the AAAI conference on artificial intelligence}, pp. \bibinfo{pages}{6892--6899}.
\bibitem[{Radford et~al.(2021)Radford, Kim, Hallacy, Ramesh, Goh, Agarwal, Sastry, Askell, Mishkin, Clark et~al.}]{clip}
\bibinfo{author}{Radford, A.}, \bibinfo{author}{Kim, J.W.}, \bibinfo{author}{Hallacy, C.}, \bibinfo{author}{Ramesh, A.}, \bibinfo{author}{Goh, G.}, \bibinfo{author}{Agarwal, S.}, \bibinfo{author}{Sastry, G.}, \bibinfo{author}{Askell, A.}, \bibinfo{author}{Mishkin, P.}, \bibinfo{author}{Clark, J.}, et~al., \bibinfo{year}{2021}.
\newblock \bibinfo{title}{Learning transferable visual models from natural language supervision}, in: \bibinfo{booktitle}{International conference on machine learning}, \bibinfo{organization}{PMLR}. pp. \bibinfo{pages}{8748--8763}.
\bibitem[{Shang et~al.(2019)Shang, Xiao, Ma, Li and Sun}]{gamenet}
\bibinfo{author}{Shang, J.}, \bibinfo{author}{Xiao, C.}, \bibinfo{author}{Ma, T.}, \bibinfo{author}{Li, H.}, \bibinfo{author}{Sun, J.}, \bibinfo{year}{2019}.
\newblock \bibinfo{title}{Gamenet: Graph augmented memory networks for recommending medication combination}, in: \bibinfo{booktitle}{Proceedings of the AAAI Conference on Artificial Intelligence}, pp. \bibinfo{pages}{1126--1133}.
\bibitem[{Tan et~al.(2024)Tan, Rong, Zhao, Bian, Xu, Huang, Cheng and Meng}]{tan2024natural}
\bibinfo{author}{Tan, J.}, \bibinfo{author}{Rong, Y.}, \bibinfo{author}{Zhao, K.}, \bibinfo{author}{Bian, T.}, \bibinfo{author}{Xu, T.}, \bibinfo{author}{Huang, J.}, \bibinfo{author}{Cheng, H.}, \bibinfo{author}{Meng, H.}, \bibinfo{year}{2024}.
\newblock \bibinfo{title}{Natural language-assisted multi-modal medication recommendation}, in: \bibinfo{booktitle}{Proceedings of the 33rd ACM International Conference on Information and Knowledge Management}, pp. \bibinfo{pages}{2200--2209}.
\bibitem[{Tran et~al.(2017)Tran, Liu, Zhou and Jin}]{missing_1}
\bibinfo{author}{Tran, L.}, \bibinfo{author}{Liu, X.}, \bibinfo{author}{Zhou, J.}, \bibinfo{author}{Jin, R.}, \bibinfo{year}{2017}.
\newblock \bibinfo{title}{Missing modalities imputation via cascaded residual autoencoder}, in: \bibinfo{booktitle}{Proceedings of the IEEE conference on computer vision and pattern recognition}, pp. \bibinfo{pages}{1405--1414}.
\bibitem[{Wang et~al.(2017)Wang, Liu, Liu, Wang, Long and Qian}]{earlywork1}
\bibinfo{author}{Wang, M.}, \bibinfo{author}{Liu, M.}, \bibinfo{author}{Liu, J.}, \bibinfo{author}{Wang, S.}, \bibinfo{author}{Long, G.}, \bibinfo{author}{Qian, B.}, \bibinfo{year}{2017}.
\newblock \bibinfo{title}{Safe medicine recommendation via medical knowledge graph embedding. arxiv}.
\newblock \bibinfo{journal}{Information Retrieval} .
\bibitem[{Wang et~al.(2021)Wang, Wei, Yin, Wu, Song and Nie}]{dualgnn}
\bibinfo{author}{Wang, Q.}, \bibinfo{author}{Wei, Y.}, \bibinfo{author}{Yin, J.}, \bibinfo{author}{Wu, J.}, \bibinfo{author}{Song, X.}, \bibinfo{author}{Nie, L.}, \bibinfo{year}{2021}.
\newblock \bibinfo{title}{Dualgnn: Dual graph neural network for multimedia recommendation}.
\newblock \bibinfo{journal}{IEEE Transactions on Multimedia} \bibinfo{volume}{25}, \bibinfo{pages}{1074--1084}.
\bibitem[{Wang et~al.(2015)Wang, Arora, Livescu and Bilmes}]{deep}
\bibinfo{author}{Wang, W.}, \bibinfo{author}{Arora, R.}, \bibinfo{author}{Livescu, K.}, \bibinfo{author}{Bilmes, J.}, \bibinfo{year}{2015}.
\newblock \bibinfo{title}{On deep multi-view representation learning}, in: \bibinfo{booktitle}{International conference on machine learning}, \bibinfo{organization}{PMLR}. pp. \bibinfo{pages}{1083--1092}.
\bibitem[{Wang et~al.(2022)Wang, Liang and Liu}]{earlywork3}
\bibinfo{author}{Wang, Z.}, \bibinfo{author}{Liang, Y.}, \bibinfo{author}{Liu, Z.}, \bibinfo{year}{2022}.
\newblock \bibinfo{title}{Ffbdnet: Feature fusion and bipartite decision networks for recommending medication combination}, in: \bibinfo{booktitle}{Joint European Conference on Machine Learning and Knowledge Discovery in Databases}, \bibinfo{organization}{Springer}. pp. \bibinfo{pages}{419--436}.
\bibitem[{Wu et~al.(2023)Wu, He, Mao, Li and Cambria}]{wu2023megacare}
\bibinfo{author}{Wu, J.}, \bibinfo{author}{He, K.}, \bibinfo{author}{Mao, R.}, \bibinfo{author}{Li, C.}, \bibinfo{author}{Cambria, E.}, \bibinfo{year}{2023}.
\newblock \bibinfo{title}{Megacare: Knowledge-guided multi-view hypergraph predictive framework for healthcare}.
\newblock \bibinfo{journal}{Information Fusion} \bibinfo{volume}{100}, \bibinfo{pages}{101939}.
\bibitem[{Wu et~al.(2022)Wu, Qiu, Jiang, Qi and Wu}]{cognet}
\bibinfo{author}{Wu, R.}, \bibinfo{author}{Qiu, Z.}, \bibinfo{author}{Jiang, J.}, \bibinfo{author}{Qi, G.}, \bibinfo{author}{Wu, X.}, \bibinfo{year}{2022}.
\newblock \bibinfo{title}{Conditional generation net for medication recommendation}, in: \bibinfo{booktitle}{Proceedings of the ACM Web Conference 2022}, pp. \bibinfo{pages}{935--945}.
\bibitem[{Xu et~al.(2018)Xu, Hu, Leskovec and Jegelka}]{gin}
\bibinfo{author}{Xu, K.}, \bibinfo{author}{Hu, W.}, \bibinfo{author}{Leskovec, J.}, \bibinfo{author}{Jegelka, S.}, \bibinfo{year}{2018}.
\newblock \bibinfo{title}{How powerful are graph neural networks?}
\newblock \bibinfo{journal}{arXiv preprint arXiv:1810.00826} .
\bibitem[{Yang et~al.(2021a)Yang, Xiao, Glass and Sun}]{micron}
\bibinfo{author}{Yang, C.}, \bibinfo{author}{Xiao, C.}, \bibinfo{author}{Glass, L.}, \bibinfo{author}{Sun, J.}, \bibinfo{year}{2021}a.
\newblock \bibinfo{title}{Change matters: Medication change prediction with recurrent residual networks}, in: \bibinfo{booktitle}{Proceedings of the Thirtieth International Joint Conference on Artificial Intelligence 2021}, pp. \bibinfo{pages}{3728--3734}.
\bibitem[{Yang et~al.(2021b)Yang, Xiao, Ma, Glass and Sun}]{safedrug}
\bibinfo{author}{Yang, C.}, \bibinfo{author}{Xiao, C.}, \bibinfo{author}{Ma, F.}, \bibinfo{author}{Glass, L.}, \bibinfo{author}{Sun, J.}, \bibinfo{year}{2021}b.
\newblock \bibinfo{title}{Safedrug: Dual molecular graph encoders for safe drug recommendations}, in: \bibinfo{booktitle}{Proceedings of the Thirtieth International Joint Conference on Artificial Intelligence, {IJCAI} 2021}, pp. \bibinfo{pages}{3735--3741}.
\bibitem[{Yang et~al.(2023)Yang, Zeng, Wu and Yan}]{molerec}
\bibinfo{author}{Yang, N.}, \bibinfo{author}{Zeng, K.}, \bibinfo{author}{Wu, Q.}, \bibinfo{author}{Yan, J.}, \bibinfo{year}{2023}.
\newblock \bibinfo{title}{Molerec: Combinatorial drug recommendation with substructure-aware molecular representation learning}, in: \bibinfo{booktitle}{Proceedings of the ACM Web Conference 2023}, pp. \bibinfo{pages}{4075--4085}.
\bibitem[{Yi et~al.(2022)Yi, Wang, Ounis and Macdonald}]{mmgcl}
\bibinfo{author}{Yi, Z.}, \bibinfo{author}{Wang, X.}, \bibinfo{author}{Ounis, I.}, \bibinfo{author}{Macdonald, C.}, \bibinfo{year}{2022}.
\newblock \bibinfo{title}{Multi-modal graph contrastive learning for micro-video recommendation}, in: \bibinfo{booktitle}{Proceedings of the 45th International ACM SIGIR Conference on Research and Development in Information Retrieval}, pp. \bibinfo{pages}{1807--1811}.
\bibitem[{Zhang et~al.(2022)Zhang, Liu, Zhou, Miao, Wang and Tang}]{traRSs2}
\bibinfo{author}{Zhang, L.}, \bibinfo{author}{Liu, Y.}, \bibinfo{author}{Zhou, X.}, \bibinfo{author}{Miao, C.}, \bibinfo{author}{Wang, G.}, \bibinfo{author}{Tang, H.}, \bibinfo{year}{2022}.
\newblock \bibinfo{title}{Diffusion-based graph contrastive learning for recommendation with implicit feedback}, in: \bibinfo{booktitle}{International Conference on Database Systems for Advanced Applications}, \bibinfo{organization}{Springer}. pp. \bibinfo{pages}{232--247}.
\bibitem[{Zhang et~al.(2017)Zhang, Chen, Tang, Stewart and Sun}]{leap}
\bibinfo{author}{Zhang, Y.}, \bibinfo{author}{Chen, R.}, \bibinfo{author}{Tang, J.}, \bibinfo{author}{Stewart, W.F.}, \bibinfo{author}{Sun, J.}, \bibinfo{year}{2017}.
\newblock \bibinfo{title}{Leap: learning to prescribe effective and safe treatment combinations for multimorbidity}, in: \bibinfo{booktitle}{Proceedings of the 23rd ACM SIGKDD International Conference on Knowledge Discovery and Data Mining}, pp. \bibinfo{pages}{1315--1324}.
\bibitem[{Zhao et~al.(2021)Zhao, Li and Jin}]{missing_2}
\bibinfo{author}{Zhao, J.}, \bibinfo{author}{Li, R.}, \bibinfo{author}{Jin, Q.}, \bibinfo{year}{2021}.
\newblock \bibinfo{title}{Missing modality imagination network for emotion recognition with uncertain missing modalities}, in: \bibinfo{booktitle}{Proceedings of the 59th Annual Meeting of the Association for Computational Linguistics and the 11th International Joint Conference on Natural Language Processing (Volume 1: Long Papers)}, pp. \bibinfo{pages}{2608--2618}.
\bibitem[{Zhao et~al.(2024)Zhao, Jing, Feng, Wu, Gao and He}]{zhao2024leave}
\bibinfo{author}{Zhao, Z.}, \bibinfo{author}{Jing, Y.}, \bibinfo{author}{Feng, F.}, \bibinfo{author}{Wu, J.}, \bibinfo{author}{Gao, C.}, \bibinfo{author}{He, X.}, \bibinfo{year}{2024}.
\newblock \bibinfo{title}{Leave no patient behind: Enhancing medication recommendation for rare disease patients}, in: \bibinfo{booktitle}{Proceedings of the 47th International ACM SIGIR Conference on Research and Development in Information Retrieval}, pp. \bibinfo{pages}{533--542}.
\bibitem[{Zheng et~al.(2021)Zheng, Wang, Xu, Shen, Qin, Huai, Liu and Chen}]{longitudinal3}
\bibinfo{author}{Zheng, Z.}, \bibinfo{author}{Wang, C.}, \bibinfo{author}{Xu, T.}, \bibinfo{author}{Shen, D.}, \bibinfo{author}{Qin, P.}, \bibinfo{author}{Huai, B.}, \bibinfo{author}{Liu, T.}, \bibinfo{author}{Chen, E.}, \bibinfo{year}{2021}.
\newblock \bibinfo{title}{Drug package recommendation via interaction-aware graph induction}, in: \bibinfo{booktitle}{Proceedings of the Web Conference 2021}, pp. \bibinfo{pages}{1284--1295}.
\bibitem[{Zhou et~al.(2023a)Zhou, Zhou, Zhang and Shen}]{dragon}
\bibinfo{author}{Zhou, H.}, \bibinfo{author}{Zhou, X.}, \bibinfo{author}{Zhang, L.}, \bibinfo{author}{Shen, Z.}, \bibinfo{year}{2023}a.
\newblock \bibinfo{title}{Enhancing dyadic relations with homogeneous graphs for multimodal recommendation}, in: \bibinfo{booktitle}{ECAI 2023}. \bibinfo{publisher}{IOS Press}, pp. \bibinfo{pages}{3123--3130}.
\bibitem[{Zhou et~al.(2023b)Zhou, Lin, Liu and Miao}]{traRSs3}
\bibinfo{author}{Zhou, X.}, \bibinfo{author}{Lin, D.}, \bibinfo{author}{Liu, Y.}, \bibinfo{author}{Miao, C.}, \bibinfo{year}{2023}b.
\newblock \bibinfo{title}{Layer-refined graph convolutional networks for recommendation}, in: \bibinfo{booktitle}{2023 IEEE 39th International Conference on Data Engineering (ICDE)}, \bibinfo{organization}{IEEE}. pp. \bibinfo{pages}{1247--1259}.
\bibitem[{Zhou et~al.(2023c)Zhou, Sun, Liu, Zhang and Miao}]{traRSs4}
\bibinfo{author}{Zhou, X.}, \bibinfo{author}{Sun, A.}, \bibinfo{author}{Liu, Y.}, \bibinfo{author}{Zhang, J.}, \bibinfo{author}{Miao, C.}, \bibinfo{year}{2023}c.
\newblock \bibinfo{title}{Selfcf: A simple framework for self-supervised collaborative filtering}.
\newblock \bibinfo{journal}{ACM Transactions on Recommender Systems} \bibinfo{volume}{1}, \bibinfo{pages}{1--25}.

\end{thebibliography}



\end{document}